\documentclass{article} %
\usepackage{template}

\usepackage{amsmath,amsfonts,bm}

\def\eqref#1{equation~\ref{#1}}

\def\1{\bm{1}}

\DeclareMathAlphabet{\mathsfit}{\encodingdefault}{\sfdefault}{m}{sl}
\SetMathAlphabet{\mathsfit}{bold}{\encodingdefault}{\sfdefault}{bx}{n}

\usepackage[utf8]{inputenc} %
\usepackage[T1]{fontenc}    %
\usepackage[backref=section]{hyperref}
\usepackage{url}

\usepackage{booktabs}
\usepackage{multirow}
\usepackage{colortbl}
\usepackage{multicol}
\usepackage{amsfonts}
\usepackage{nicefrac}
\usepackage{microtype}
\usepackage[most]{tcolorbox}
\usepackage[dvipsnames]{xcolor}
\usepackage{latexsym}
\usepackage{graphicx}
\usepackage{float}
\usepackage{subcaption}
\usepackage{wrapfig}
\usepackage{dirtytalk}
\usepackage{enumitem}
\usepackage{makecell}
\usepackage{titletoc}
\usepackage{minted}
\usepackage{todonotes}

\usepackage[toc,page,header]{appendix}

\usepackage{bm}

\usepackage{tabularx} 
\usepackage{arydshln}
\usepackage{ragged2e} 
\newcolumntype{L}{>{\RaggedRight\hangafter=1\hangindent=0em}X}

\usepackage{amsmath}
\usepackage{amssymb}
\usepackage{mathtools}
\usepackage{amsthm}
\usepackage{algorithm}
\usepackage{algpseudocode}

\hypersetup{
    colorlinks=true,
    linkcolor=Maroon,
    citecolor=Cyan,
    filecolor=magenta,      
    urlcolor=magenta,
}

\usepackage[capitalize,noabbrev]{cleveref}
\crefname{section}{§}{§§}
\Crefname{section}{§}{§§}
\crefname{lemma}{lemma}{lemma}
\Crefname{lemma}{Lemma}{Lemma}

\usepackage{calligra}
\DeclareMathAlphabet{\mathcalligra}{T1}{calligra}{m}{n}

\usepackage{pifont}

\theoremstyle{plain}

\theoremstyle{definition}

\theoremstyle{remark}

\renewcommand{\paragraph}[1]{\vspace{1mm}\noindent\textbf{#1}}

\DeclareCaptionLabelFormat{cont}{#1~#2\alph{ContinuedFloat}}

\tcbset{
  promptbox/.style={
    top=10pt,
    colback=lightgray!20,
    colframe=Black,
    colbacktitle=NavyBlue,
    enhanced,
    center,
    attach boxed title to top center={yshift=-0.1in,xshift=0.0in},
    boxed title style={boxrule=0pt,colframe=white,},
  }
}
\newtcolorbox{promptbox}[2][]{promptbox, title=#2,#1}
\tcbset{
  takeawaybox/.style={
    top=10pt,
    colback=lightgray!20,
    colframe=Black,
    colbacktitle=BurntOrange,
    enhanced,
    center,
    attach boxed title to top center={yshift=-0.1in,xshift=0.0in},
    boxed title style={boxrule=0pt,colframe=white,},
  }
}
\newtcolorbox{takeawaybox}[2][]{takeawaybox, title=#2,#1}
\tcbset{
  observationbox/.style={
    top=10pt,
    colback=lightgray!20,
    colframe=Black,
    colbacktitle=YellowGreen,
    enhanced,
    center,
    attach boxed title to top center={yshift=-0.1in,xshift=0.0in},
    boxed title style={boxrule=0pt,colframe=white,},
  }
}
\newtcolorbox{observationbox}[2][]{observationbox, title=#2,#1}

\newtcbtheorem[number within=section]{prompt}{Prompt}{
    top=10pt,
    colback=lightgray!10,
    colframe=Black,
    colbacktitle=Cyan,
    coltitle=white,
    fonttitle=\bfseries,
    enhanced,
    center,
    breakable,
    separator sign={:},
    description delimiters={}{},
    boxrule=0.8pt,
    arc=3pt,
    outer arc=3pt,
    boxsep=0pt,
    left=12pt,right=12pt,top=10pt,bottom=10pt,
    lefttitle=12pt,righttitle=12pt,toptitle=5pt,bottomtitle=5pt,
    before skip=12pt,after skip=12pt,
    pad at break*=3mm,
    fontupper=\normalsize
}{pt}

\newtcbtheorem[number within=section]{tool}{Tool Card}{
    top=10pt,
    colback=lightgray!20,
    colframe=Black,
    colbacktitle=YellowGreen,
    enhanced,
    center,
    attach boxed title to top center={yshift=-0.1in,xshift=0.0in},
    boxed title style={boxrule=0pt,colframe=white,},
    breakable
}{tl} 

\lstdefinelanguage{yaml}{
  keywords={true,false,null,y,n},
  keywordstyle=\color{blue}\bfseries,
  basicstyle=\ttfamily\small,
  commentstyle=\color{gray}\itshape,
  stringstyle=\color{purple},
  moredelim=[l][\color{orange}]{\&},
  moredelim=[l][\color{magenta}]{\*},
  comment=[l]{\#},
  morecomment=[s]{/*}{*/},
  morestring=[b]',
  morestring=[b]"
} 

\definecolor{ageCaseHeader}{HTML}{4B4B4B}
\definecolor{ageTaskGray}{HTML}{F0F0F0}
\definecolor{ageTeacherBg}{HTML}{FFF0E6}
\definecolor{ageStudentBg}{HTML}{D9FBDD}
\definecolor{ageAnalysisBg}{HTML}{F0FFF2}
\definecolor{ageAnalysisGreen}{HTML}{15962B}

\newtcbtheorem[number within=section]{case}{Case}{
    top=10pt,
    enhanced,
    breakable,
    colback=white,
    colframe=ageCaseHeader,
    colbacktitle=ageCaseHeader,
    coltitle=white,
    fonttitle=\bfseries,
    separator sign={:},
    description delimiters={}{},
    boxrule=0.8pt,
    arc=3pt,
    outer arc=3pt,
    boxsep=0pt,
    left=12pt,right=12pt,top=10pt,bottom=10pt,
    lefttitle=12pt,righttitle=12pt,toptitle=5pt,bottomtitle=5pt,
    before skip=12pt,after skip=12pt,
    pad at break*=3mm,
    fontupper=\normalsize
}{cs}

\newtcolorbox{agetaskbox}{
    colback=ageTaskGray,
    colframe=ageTaskGray,
    boxrule=0pt,
    arc=0mm,
    left=2.5mm,
    right=2.5mm,
    top=1.5mm,
    bottom=1.5mm
}

\newtcolorbox{ageanalysisbox}{
    colback=ageAnalysisBg,
    colframe=ageAnalysisGreen,
    colbacktitle=ageAnalysisBg,
    coltitle=ageAnalysisGreen,
    fonttitle=\bfseries,
    title={Analysis},
    boxrule=0.7pt,
    arc=2mm,
    left=3mm,
    right=3mm,
    top=2mm,
    bottom=2mm
} 

\newcommand{\agePromptLine}[1]{\noindent\detokenize{#1}\par}
\newcommand{\agePromptBlank}{\vspace{0.55em}}

\usepackage{xspace}
\newcommand{\name}[0]{\textsc{{STI-OPD}}\xspace}
\newcommand{\drop}[1]{\textcolor{red!50!black}{$\downarrow${\scriptsize #1}}}
\newcommand{\up}[1]{\textcolor{green!50!black}{$\uparrow${\scriptsize #1}}}

\definecolor{carnelian}{rgb}{0.7, 0.11, 0.11}

\title{Stochastic Teacher Intervention for Agentic On-Policy Distillation}

\author{\textbf{Junnan Liu}$^{1}$,
\textbf{Linhao Luo}$^{1}$,
\textbf{Zhijun Chen}$^{2}$,
\textbf{Qianren Mao}$^{3}$, \\
\textbf{Thuy-Trang Vu}$^{1}$
\& \textbf{Gholamreza Haffari}$^{1}$ \\[0.4em]
\normalfont
$^{1}$Department of Data Science and AI, Faculty of Information Technology, \\
Monash University, Australia \\
$^{2}$The Hong Kong Polytechnic University, Hong Kong SAR, China \\
$^{3}$Zhongguancun Laboratory, Beijing, P.R. China
}

\date{\today}
\correspondence{Junnan Liu at
  \href{mailto:junnan.liu@monash.edu}{\texttt{junnan.liu@monash.edu}}}

\begin{document}

\maketitle
\begin{abstract}
On-policy distillation (OPD) efficiently transfers capabilities from a stronger teacher to a student language model through dense token-level supervision on student-generated rollouts and has shown promise on complex tasks such as mathematical reasoning.
However, in multi-turn agentic tasks, student decisions shape subsequent observations, causing early errors to accumulate across turns.
The resulting trajectories can drift away from the teacher's rollout distribution, making the teacher's token-level supervision less reliable or even counterproductive for OPD training.
To address this issue, we introduce \name, a stochastic teacher intervention framework for multi-turn agentic OPD.
During multi-turn interaction, \name uses teacher intervention guided by teacher--student policy discrepancy to replace the student's proposed action with a teacher-generated one to maximize the acquisition of reliable supervision.
We further develop a stochastic intervention strategy, addressing the limitations of previous threshold-based or fixed-schedule approaches, that estimates policy discrepancy using KL divergence and maps it to an intervention probability. 
By sampling whether to intervene from this probability, \name adaptively balances teacher control with student exploration.
To learn from the resulting mixed-policy trajectories, we introduce an Importance-Weighted Reverse KL objective that corrects the token sampling mismatch between teacher-generated responses and the student policy to preserve the original OPD objective.
Across tool-integrated reasoning and long-horizon interaction, \name outperforms the strongest prior OPD baseline on every evaluated benchmark and student size.
Ablations further show that both discrepancy-guided intervention and importance weighting contribute to these gains.
\end{abstract}

\section{Introduction}
\label{sec:introduction} 

\begin{wrapfigure}{l}{0.6\textwidth}
    \centering
    \includegraphics[width=.6\textwidth]{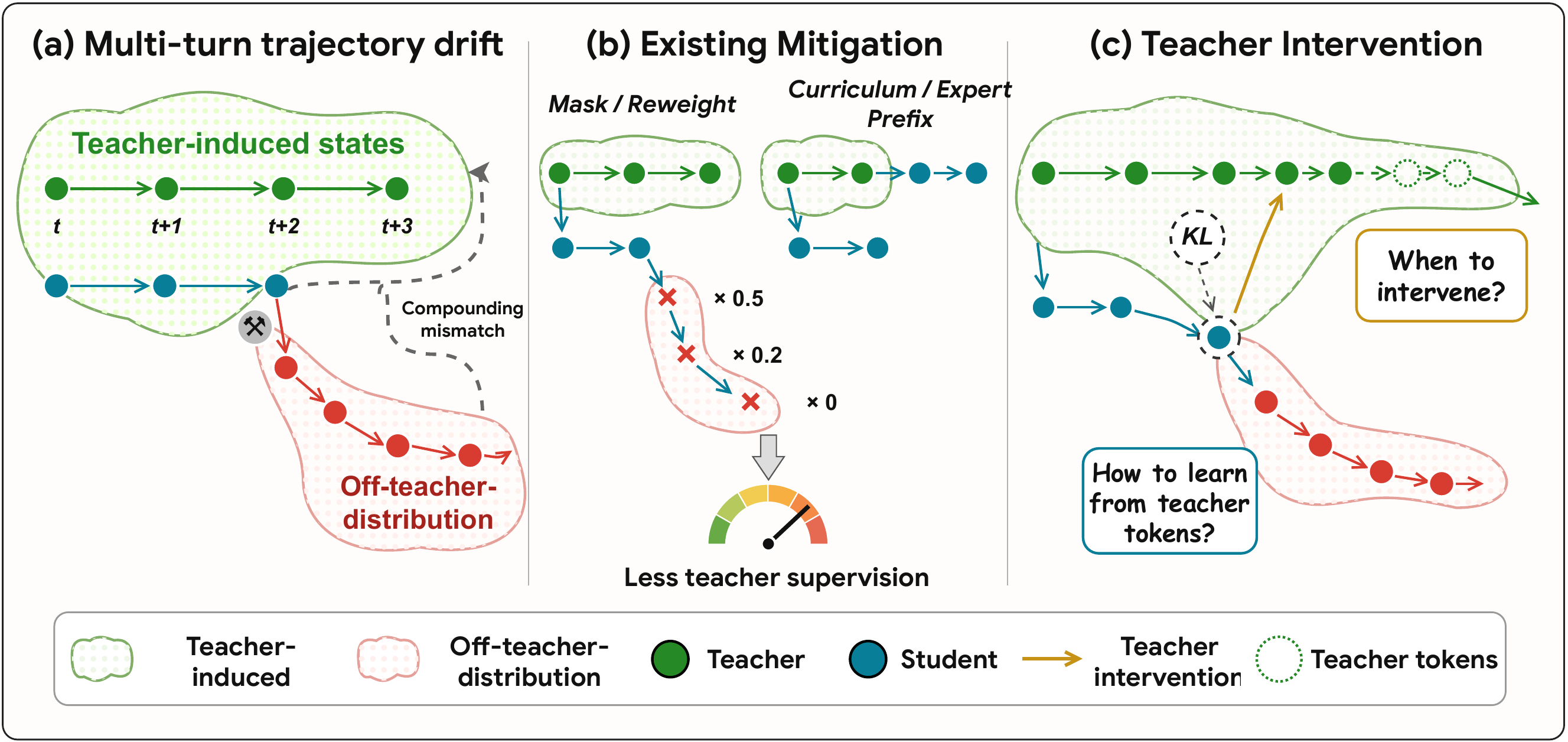}
    \caption{Illustration of trajectory-level distribution shift and turn-level teacher intervention in on-policy distillation.}
    \label{fig:intro}
    \vspace{-1.0em}
\end{wrapfigure}

On-policy distillation (OPD)~\citep{AgarwalVZSGGB24,Gu0WH24,abs-2505-09388,abs-2512-02556,lu2025onpolicydistillation,abs-2601-18734} transfers capabilities from a stronger teacher to a student large language model (LLM) by minimizing token-level KL divergence on student-generated rollouts. By supervising contexts induced by the student's own policy, OPD reduces the train--inference mismatch and provides dense token-level guidance, in contrast to the typically sparse outcome-level rewards of reinforcement learning with verifiable rewards (RLVR), leading to promising results on complex tasks such as mathematical reasoning~\citep{lu2025onpolicydistillation,abs-2603-07079}. Extending OPD to multi-turn agentic tasks~\citep{abs-2409-12122,ShridharYCBTH21}, however, introduces a trajectory drift challenge: because each action shapes subsequent observations, an incorrect tool call, or even a valid action unlikely under the teacher policy, can lead the student into \emph{off-teacher-distribution states} rarely encountered in teacher-generated rollouts~\citep{abs-2605-07725,abs-2604-24005}, as illustrated in \Cref{fig:intro}. As such deviations accumulate across turns, teacher--student policy discrepancy can increase, making the teacher's token-level guidance on the resulting histories less reliable or even counterproductive~\citep{abs-2604-13016}. Consequently, supervising student-generated tokens alone may be insufficient to control the distribution shift induced by multi-turn interaction.

Existing methods address this instability through two strategies.
One masks or reweights the distillation loss using signals derived from KL divergence~\citep{abs-2605-07725,abs-2606-19659} or model confidence~\citep{abs-2604-10688}.
The other controls training-trajectory construction by progressively adjusting rollout depth~\citep{abs-2604-24005} or mixing teacher- and student-generated turns through a curriculum~\citep{abs-2606-15912}.
However, loss filtering will lose supervision signals and cannot directly correct the action that determines the next environment transition.
Predefined curricula and rollout-depth control can prevent the student's exposure to unreliable supervision, but do not adaptively assess whether its current response requires correction.
Recent work has further explored \emph{teacher intervention} that allows the teacher to take over from the students' generation at specific points~\citep{abs-2607-26057}. 
By replacing the student's proposed action with a teacher-generated one, these methods could limit the accumulation of trajectory drift and provide more reliable supervision on subsequent turns.

However, making teacher intervention effective in multi-turn OPD still requires addressing two key challenges.
First, \emph{\textbf{when should the teacher intervene?}}
Predefined intervention schedules fail to account for immediate changes in teacher--student policy discrepancy during multi-turn interactions~\citep{abs-2606-15912}. 
On the other hand, recent adaptive approaches use discrepancy estimates to guide intervention through decision boundaries, hysteretic thresholds, or lookahead validation~\citep{abs-2605-19433,abs-2607-29078,abs-2608-01953}. 
However, these estimates are subject to statistical noise, and threshold-based decisions can amplify estimation errors into abrupt changes in teacher control.
A poor intervention strategy may therefore miss opportunities to correct trajectory drift or intervene unnecessarily, increasing teacher dependence and computational cost. 
Second, \emph{\textbf{how should the student learn from teacher-generated tokens?}}
Teacher intervention introduces tokens sampled from the teacher policy into training, whereas standard OPD defines its reverse-KL objective under student-generated tokens.
Existing intervention-based OPD methods handle these teacher-generated tokens using different surrogate objectives~\citep{abs-2606-15912,abs-2607-26057,abs-2608-01953}.
Some adopt forward-KL or SFT-style objectives, which depart from the reverse-KL formulation of standard OPD and may affect the retention of previously acquired capabilities.
Others directly apply a sampled reverse-KL objective to teacher-generated tokens.
However, because these tokens are sampled from the teacher rather than the student policy, directly evaluating the standard reverse-KL integrand does not recover its intended expectation under the student policy.
Thus, learning from teacher intervention requires correcting this token-sampling mismatch while preserving the conditional reverse-KL objective of OPD.

To address these challenges, we propose \name, a discrepancy-guided \underline{S}tochastic \underline{T}eacher \underline{I}ntervention framework for multi-turn agentic OPD.
We first develop a stochastic intervention strategy that samples whether to intervene based on teacher--student policy discrepancy. 
Specifically, \name uses KL divergence estimated by $k_3$ estimator~\citep{Schulman2020KL} to measure the divergence between the teacher and student policies at each turn, and maps this estimate into an intervention probability.
This probability increases smoothly with the discrepancy, favoring teacher intervention when the policies diverge while preserving student exploration, without a hand-tuned threshold or fixed intervention budget.
To learn from the resulting mixed-policy trajectories, we further introduce an importance-weighted reverse KL objective that uses importance sampling~\citep{SchulmanWDRK17} to correct the token sampling mismatch between teacher-generated responses and the student policy while preserving the original OPD objective formulation.
Intervention thus controls which histories are visited, while importance weighting corrects the token-sampling mismatch introduced by teacher generation. 

Extensive experiments are conducted in two multi-turn agentic settings, spanning tool-integrated reasoning~\citep{aime24,aime25,aime26,abs-2605-00674,abs-2311-12022,JainHGLYZWSSS25} and long-horizon agentic interaction~\citep{ShridharYCBTH21,WangJCA22}.
\name outperforms all distillation baselines on each benchmark and at both student scales, with its largest margins on long-horizon tasks, where compounding error is most severe. 
Ablations show that different components of \name contribute to these gains and further analysis reveals the mechanism, scalability, and efficiency of \name.

\section{Related Work}
\label{sec:related_work}

\vspace{-0.5em}
\paragraph{On-policy Distillation for Multi-Turn Agents.}
Small errors compound across turns and move student rollouts into histories on which teacher supervision is unreliable~\citep{abs-2605-07725,abs-2604-24005,abs-2606-15912}.
Existing methods manage this instability by controlling either the supervision signal or the rollout distribution.
The first group reweights or masks the distillation loss: SCOPE~\citep{abs-2604-10688} calibrates supervision using trajectory correctness together with teacher or student confidence, and SAGE-OPD~\citep{abs-2606-19659} applies guidance selectively within multi-turn rollouts.
The second group reshapes which trajectories the student is trained on: TCOD~\citep{abs-2604-24005} progressively expands the trajectory depth exposed to the student, Guided-OPD~\citep{abs-2606-15912} mixes teacher- and student-generated turns by a curriculum, TurnOPD~\citep{abs-2607-05804} allocates rollout depth and loss weight across turns, and ReOPD~\citep{abs-2607-04763} replays teacher trajectories as prefixes, trading student occupancy against teacher reliability.
Closest to \name, Relay-OPD~\citep{abs-2607-26057} temporarily hands generation to the teacher after detecting a failed prefix, and FutureBridge-OPD~\citep{abs-2608-01953} scores a short teacher bridge by its effect on the subsequent student continuation.
Methods in the second group that hand control to the teacher face an estimation problem that neither group addresses: the tokens produced during intervention are sampled from the teacher, yet they are supervised with an integrand that assumes student sampling.

\vspace{-0.5em}
\paragraph{Expert Intervention and Distribution Shift.}
Teacher intervention is an instance of interactive imitation learning, where the learner's own actions change its future state distribution.
DAgger repeatedly queries an expert on states induced by the learner~\citep{RossGB11}, scheduled sampling gradually replaces ground-truth prefixes with model-generated prefixes in autoregressive prediction~\citep{BengioVJS15}, and more recent methods treat expert interventions themselves as learning signals~\citep{LuoDZ0L24}.
The same idea has been carried into LLM distillation: MOTAB monitors student trajectories, backtracks when they cross an adaptive boundary, and invokes the teacher from the last safe state~\citep{abs-2605-19433}, while SMRC-SD applies privileged reference guidance to multi-turn agents only when the student's execution state matches a supported reference state~\citep{abs-2608-05219}.
The intervention rules closest to ours are all specified by hand, making them inflexible and sensitive to noisy turn-level estimates: Guided-OPD follows a predefined curriculum~\citep{abs-2606-15912}, Relay-OPD uses a continuation-asymmetry trigger~\citep{abs-2607-26057}, FutureBridge-OPD validates guidance through future rollouts~\citep{abs-2608-01953}, and DASH-OPD switches executors on hysteretic thresholds over accumulated log-probability-ratio evidence~\citep{abs-2607-29078}.
\name differs in both respects: each decision is stochastic, with a probability read off the instantaneous teacher--student KL divergence rather than a threshold, and the estimation mismatch that intervention introduces is corrected by importance sampling so that teacher-generated tokens still serve the target OPD objective.

\section{Methodology}
\label{sec:methodology}

\begin{figure}[t]
    \centering
    \includegraphics[width=\linewidth]{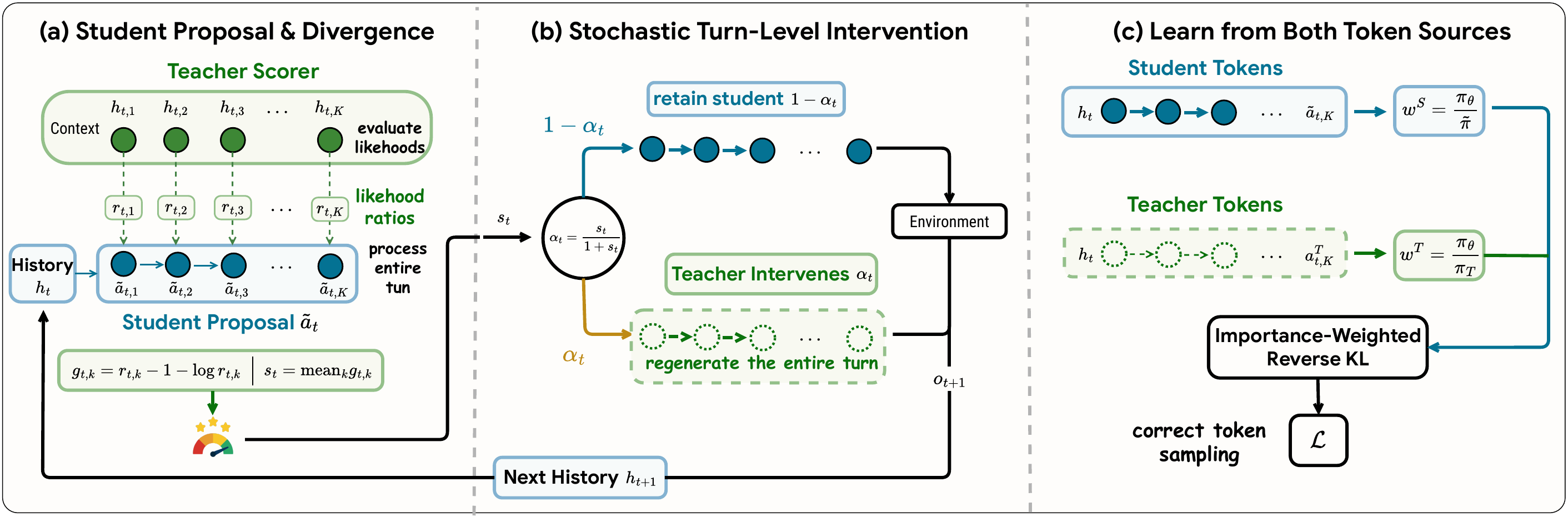}
    \caption{Overview of \name. A non-negative $k_3$ discrepancy estimate determines whether to retain the student proposal or replace it with a teacher-generated response before execution. The executed response is used for training, with importance ratios recovering the conditional reverse-KL expectation for teacher-generated tokens.}
    \label{fig:framework_overview}
    \vspace{-2.0em}
\end{figure}

\vspace{-0.5em}
As illustrated in \Cref{fig:framework_overview}, \name first estimates teacher--student policy discrepancy from a student-proposed response using the $k_3$ estimator.
The resulting score determines the probability of retaining the proposal or replacing it with a teacher-regenerated response.
The selected response is executed to update the interaction history and added to the training buffer.
During optimization, an importance ratio corrects the sampling distribution of teacher-generated tokens to account for the generating policy, recovering the conditional reverse-KL expectation for teacher-generated tokens.
\vspace{-0.5em}

\subsection{Problem Formulation}
\label{sec:problem_formulation} 

\vspace{-0.5em}
\paragraph{Multi-Turn Agentic Task.}
Given a query $q\sim\mathcal{D}$, an agent interacts with an environment over multiple turns to solve the task~\citep{YaoZYDSN023,ShridharYCBTH21,WangJCA22}.
At turn $t$, the agent observes $o_t$ and samples a response $a_t\sim\pi_\theta(\cdot\mid h_t)$ from its policy $\pi_\theta$, conditioned on the interaction history $h_t=(q,o_1,a_1,\ldots,o_{t-1},a_{t-1},o_t)$.
Each response contains natural-language reasoning and possible subsequent tool calls, environment actions, and the final answer.
For each tool call or environment action, the environment executes the operation specified in $a_t$ and returns a new observation $o_{t+1}\sim P(\cdot\mid h_t,a_t)$.
This loop continues until termination, yielding a trajectory $\tau=(q,o_1,a_1,\ldots,o_H,a_H)$ of $H$ turns, with terminal feedback omitted.
We denote its distribution by $p_{\pi_\theta}(\tau\mid q)$. 
Within each turn, the response $a_t=(x_{t,1},\ldots,x_{t,L_t})$ is generated autoregressively, where $L_t$ is its length.
Given the token-level context $c_{t,k}=(h_t,x_{t,<k})$, its probability factorizes as
\begin{equation}
    \setlength\abovedisplayskip{1pt}
    \setlength\belowdisplayskip{1pt}
    \pi_\theta(a_t\mid h_t)
    =\prod_{k=1}^{L_t}\pi_\theta(x_{t,k}\mid c_{t,k}).
    \label{eq:turn_policy_factorization}
\end{equation}

\vspace{-0.5em}
\paragraph{On-Policy Distillation.}
Let $\pi_T$ be a fixed teacher policy and $\pi_\theta$ the trainable student policy.
In OPD, the teacher provides token-level supervision on trajectories generated by the student.
Compared with offline distillation from fixed expert trajectories, OPD reduces the mismatch between training contexts and those encountered by the student at inference time~\citep{AgarwalVZSGGB24,Gu0WH24,lu2025onpolicydistillation}.
Using reverse KL divergence, the ideal multi-turn OPD objective is
\begin{equation}
    \setlength\abovedisplayskip{1.3pt}
    \setlength\belowdisplayskip{1.3pt}
    \mathcal{L}_{\mathrm{OPD}}(\theta)
    =
    \mathbb{E}_{q\sim\mathcal{D},\,\tau\sim p_{\pi_\theta}(\cdot\mid q)}
    \left[
        \sum_{t=1}^{H}\sum_{k=1}^{L_t}
        D_{\mathrm{KL}}\!\left(
            \pi_\theta(\cdot\mid c_{t,k})
            \,\|\,
            \pi_T(\cdot\mid c_{t,k})
        \right)
    \right].
    \label{eq:multi_turn_opd}
\end{equation}
With a fixed context $c$, to reduce the heavy overhead of computing the full KL divergence, each reverse-KL term can adopt a sampling form $\mathbb{E}_{x\sim\pi_\theta(\cdot\mid c)} \left[ 
    \log\pi_\theta(x\mid c)
    -
    \log\pi_T(x\mid c)
\right]$.
Sampled OPD therefore relies on both student-induced contexts and student-generated tokens.

\vspace{-0.5em}
\paragraph{Policy Discrepancy along Trajectories.}
As discussed in \Cref{sec:introduction}, multi-turn interaction can drive the student toward histories rarely encountered under teacher rollouts, where teacher guidance may become less reliable.
Large policy discrepancies at these histories can hinder effective distillation~\citep{abs-2604-13016}.
We quantify the teacher--student policy discrepancy at history $h_t$ by
\begin{equation}
    \setlength\abovedisplayskip{1.3pt}
    \setlength\belowdisplayskip{1.3pt}
    \kappa(h_t)
    =
    D_{\mathrm{KL}}\!\left(
        \pi_\theta(\cdot\mid h_t)
        \,\|\,
        \pi_T(\cdot\mid h_t)
    \right),
    \label{eq:instantaneous_kl}
\end{equation}
where each policy defines a distribution over complete turn-level responses.
Because student actions shape subsequent observations, a deviation at one turn can affect all later decisions.
The resulting distribution shift may compound across turns and further increase policy discrepancy~\citep{abs-2604-24005,abs-2606-15912}.
This motivates intervening before a proposed response is executed, allowing teacher guidance to influence the next environment transition rather than only the training loss.

\subsection{Stochastic Teacher Intervention}
\label{sec:stochastic_teacher_intervention}

An effective intervention strategy should respond to the policy discrepancy at the current turn while accounting for uncertainty in its estimate.
Predefined schedules overlook immediate changes in discrepancy, while hard thresholds can turn estimation errors into abrupt changes in teacher control~\citep{abs-2606-15912,abs-2604-24005}.
Poor intervention decisions may therefore allow trajectory drift to persist or introduce unnecessary teacher intervention.
These considerations motivate combining an unbiased and low-variance discrepancy estimate with an intervention probability that varies smoothly with the estimated score.

 \vspace{-0.5em}
\paragraph{Turn-Level Teacher Intervention.}
We first specify how intervention modifies a rollout.
Let $\pi_{\bar\theta}$ denote the rollout student, which may be a lagged copy of the trainable policy $\pi_\theta$.
In the strictly on-policy setting, $\pi_{\bar\theta}=\pi_\theta$.
At turn $t$, the rollout student first proposes a complete response $\widetilde{a}_t\sim\pi_{\bar\theta}(\cdot\mid h_t)$.
An indicator $z_t\in\{0,1\}$ determines whether to retain this proposal ($z_t=0$) or replace it with a teacher response $a_t^T\sim\pi_T(\cdot\mid h_t)$ generated from the same history ($z_t=1$).
The resulting trajectory is
\begin{equation}
    \setlength\abovedisplayskip{1.3pt}
    \setlength\belowdisplayskip{1.3pt}
    \tau=(q,o_1,a_1,\ldots,o_H,a_H),
    \qquad
    a_t=
    \begin{cases}
        \widetilde{a}_t, & z_t=0,\\
        a_t^T, & z_t=1.
    \end{cases}
    \label{eq:intervened_trajectory}
\end{equation}
Each intervention replaces the entire response, keeping its reasoning and selected operation consistent.
For a tool call or environment action, the environment executes the operation specified in $a_t$ and returns $o_{t+1}\sim P(\cdot\mid h_t,a_t)$.
If the episode continues, the student resumes generation from $h_{t+1}=(h_t,a_t,o_{t+1})$.
Thus, $h_t$ consists of previously executed responses from either the student or the teacher, together with the corresponding environment observations.
Both policies condition on this same history at the current turn.
A discarded student proposal is neither appended to the history nor included in the training buffer.

An optimal intervention strategy must balance teacher guidance with autonomous student execution.
Unlike random intervention with predefined probabilities~\citep{abs-2606-15912} or switching at fixed thresholds, we propose a discrepancy-conditioned \emph{stochastic intervention strategy}.
The intervention probability varies smoothly with the estimated teacher--student discrepancy, allowing teacher guidance to adapt at each turn while preserving opportunities for student execution (\Cref{fig:framework_overview}~(b)).

\vspace{-0.5em}
\paragraph{Discrepancy Estimation.}
Following the policy discrepancy defined in \Cref{eq:instantaneous_kl}, we assess the current student proposal $\widetilde{a}_t=(\widetilde{x}_{t,1},\ldots,\widetilde{x}_{t,\widetilde{L}_t})$ using token-level reverse KL.
At each token position $k$, we measure the discrepancy between the student's and teacher's next-token distributions conditioned on $\widetilde{c}_{t,k}=(h_t,\widetilde{x}_{t,<k})$ using
$D_{\mathrm{KL}}(\pi_{\bar\theta}(\cdot\mid\widetilde{c}_{t,k})\|\pi_T(\cdot\mid\widetilde{c}_{t,k}))$. 

Exact evaluation requires summing over the full vocabulary at every position.
For efficiency, the common practice is to use the KL estimator~\citep{Schulman2020KL}, which requires only the teacher and student probabilities of each sampled token $\widetilde{x}_{t,k}$.
We define the teacher-to-student likelihood ratio as
\begin{equation}
    \setlength\abovedisplayskip{1.3pt}
    \setlength\belowdisplayskip{1.3pt}
    r_{t,k}
    =
    \frac{\pi_T(\widetilde{x}_{t,k}\mid\widetilde{c}_{t,k})}
         {\pi_{\bar\theta}(\widetilde{x}_{t,k}\mid\widetilde{c}_{t,k})}.
    \label{eq:teacher_student_ratio}
\end{equation} 
To provide an unbiased and low-variance estimate of the token-level reverse KL, we use the $k_3$ estimator~\citep{Schulman2020KL},
\begin{equation}
    \setlength\abovedisplayskip{1.3pt}
    \setlength\belowdisplayskip{1.3pt}
    g_{t,k}=r_{t,k}-1-\log r_{t,k}.
    \label{eq:k3_estimator}
\end{equation}
For $r_{t,k}>0$, $g_{t,k}$ is non-negative because $\log r_{t,k}\leq r_{t,k}-1$. 
Under common support, $g_{t,k}$ is a pointwise non-negative and unbiased
estimator of the token-level reverse KL under student sampling.
Moreover, in the near-policy regime, $k_3$ removes the first-order fluctuation present in $k_1$, yielding lower-order variance, which makes it a more stable signal for intervention.
We aggregate the token-level estimates into the length-normalized turn score to remove the direct scaling of the score with response length.
\begin{equation}
    \setlength\abovedisplayskip{1.3pt}
    \setlength\belowdisplayskip{1.3pt}
    s_t
    =
    \frac{1}{\widetilde{L}_t}
    \sum_{k=1}^{\widetilde{L}_t}g_{t,k}.
    \label{eq:normalized_turn_score}
\end{equation}
The score $s_t$ summarizes token-level discrepancy along the sampled proposal and serves as a length-normalized proxy for turn-level policy discrepancy.
It is distinct from the full response-level KL $\kappa(h_t)$ in \Cref{eq:instantaneous_kl}.

\vspace{-0.5em}
\paragraph{Discrepancy-Conditioned Intervention.}
To make intervention responsive to discrepancy without imposing a hard decision boundary, we set the intervention odds equal to $s_t$. 
In this paper, we adopt the following simple monotone mapping:
\begin{equation}
    \setlength\abovedisplayskip{1.3pt}
    \setlength\belowdisplayskip{1.3pt}
    \alpha_t=\frac{s_t}{1+s_t},
    \qquad
    z_t\sim\operatorname{Bernoulli}(\alpha_t),
    \label{eq:intervention_probability}
\end{equation}
where $z_t=1$ denotes teacher intervention.
This mapping assigns higher intervention probabilities to turns with larger estimated discrepancies, without a hand-tuned threshold.
When the teacher and student policies are close, the $k_3$-based score is second order in the likelihood-ratio deviation, so $\alpha_t$ remains close to zero to allow student execution and exploration.
When the policies diverge, $\alpha_t$ approaches one, increasing the likelihood of teacher intervention to correct the student's trajectory but still allowing  nonzero student execution probability with the stochastic Bernoulli sampling.

\subsection{Learning from Teacher Intervention}
\label{sec:learning_from_teacher_intervention}

Teacher intervention helps reduce error propagation across turns, but it produces mixed-policy trajectories in which executed responses may come from either the student or the teacher.
Standard OPD assumes student-generated trajectories, with the sampled reverse-KL loss taking an expectation over tokens drawn from $\pi_\theta(\cdot\mid c)$.
In intervened turns, however, the executed tokens are drawn from $\pi_T(\cdot\mid c)$.
Directly applying this loss to teacher-generated tokens without correction does not estimate the intended conditional reverse KL.
We therefore use importance-weighted reverse KL to correct this token-sampling mismatch and recover the expectation under the student policy (\Cref{fig:framework_overview}~(c)).

\vspace{-0.5em}
\paragraph{Importance-Weighted Reverse KL.}
For a teacher-generated token, we define the importance ratio as
\begin{equation}
    \setlength\abovedisplayskip{0.0pt}
    \setlength\belowdisplayskip{0.0pt}
    w_\theta(x,c)
    =
    \frac{\pi_\theta(x\mid c)}{\pi_T(x\mid c)}.
    \label{eq:teacher_importance_ratio}
\end{equation}
Assuming that $\pi_\theta(\cdot\mid c)$ is absolutely continuous with respect to $\pi_T(\cdot\mid c)$, importance sampling yields
\begin{equation}
    \setlength\abovedisplayskip{1.3pt}
    \setlength\belowdisplayskip{1.3pt}
\begin{split}
    &\mathbb{E}_{x\sim\pi_T(\cdot\mid c)}
    \left[
        w_\theta(x,c)
        \log\frac{\pi_\theta(x\mid c)}{\pi_T(x\mid c)}
    \right] =
    \sum_x \pi_T(x\mid c)
    \frac{\pi_\theta(x\mid c)}{\pi_T(x\mid c)}
    \log\frac{\pi_\theta(x\mid c)}{\pi_T(x\mid c)} \\
    &\quad=
    \mathbb{E}_{x\sim\pi_\theta(\cdot\mid c)}
    \left[
        \log\frac{\pi_\theta(x\mid c)}{\pi_T(x\mid c)}
    \right]
    =
    D_{\mathrm{KL}}\!\left(
        \pi_\theta(\cdot\mid c)
        \,\|\,
        \pi_T(\cdot\mid c)
    \right).
\end{split}
    \label{eq:importance_weighted_reverse_kl}
\end{equation}
Teacher-generated tokens can therefore estimate the same conditional reverse-KL term once their token-sampling distribution is corrected.

\vspace{-0.5em}
\paragraph{Combined Training Objective.}
Once the task is complete, the full executed trajectory is added to the training buffer.
Consequently, each turn contributes either the retained student proposal when $z_t=0$ or the regenerated teacher response when $z_t=1$.
For a retained student token, no sampling correction is required because
$\widetilde{x}_{t,k}$ is already sampled from $\pi_\theta$.
For a teacher-generated token, we define the importance ratio as $\omega^{T}_{t,k}(\theta) = \pi_\theta(x^{T}_{t,k}\mid c^{T}_{t,k}) /\pi_T(x^{T}_{t,k}\mid c^{T}_{t,k})$.
Following the standard OPD formulation based on sampled tokens, the optimization objective is the empirical mean of the sampled reverse-KL terms over all generated tokens:
\begin{equation}
    \setlength\abovedisplayskip{1.3pt}
    \setlength\belowdisplayskip{1.3pt}
    \widehat{\mathcal{L}}_{\name}(\theta)
    =
    \frac{1}{N_{\mathrm{tok}}}
    \sum_{t=1}^{H}
    \left[
        (1-z_t)
        \sum_{k=1}^{\widetilde{L}_t}
        \ell_\theta(
            \widetilde{x}_{t,k},
            \widetilde{c}_{t,k}
        )
        +
        z_t
        \sum_{k=1}^{L_t^T}
        \omega^T_{t,k}(\theta)
        \ell_\theta(
            x^T_{t,k},
            c^T_{t,k}
        )
    \right],
    \label{eq:intervention_corrected_objective}
\end{equation}
where $\ell_\theta(x,c)=\log\pi_\theta(x\mid c)-\log\pi_T(x\mid c)$ is the token-level student-to-teacher log-ratio, whose expectation under $x\sim\pi_\theta(\cdot\mid c)$ equals the conditional reverse KL, and $N_{\mathrm{tok}}=\sum_{t=1}^{H}((1-z_t)\widetilde{L}_t+z_tL_t^T)$ is the number of optimization tokens in the episode.  
For retained student turns, the first term is exactly the standard sampled OPD objective because the tokens are drawn from $\pi_\theta$.
For intervened turns, the second term uses importance weighting to correct the change in token-sampling distribution from $\pi_T$ to $\pi_\theta$.
By \Cref{eq:importance_weighted_reverse_kl}, both branches therefore estimate the same conditional reverse-KL objective at their respective sampled contexts. 
Importantly, the identity in \Cref{eq:importance_weighted_reverse_kl} holds at a fixed context.
The importance correction accounts for the token-sampling distribution of the teacher branch, but does not undo the adaptive intervention decision or correct the distribution over contexts induced by intervention, including teacher-generated prefixes and subsequent interaction histories.

\section{Experiments}
\label{sec:experiments}

\subsection{Experimental Setup}
\label{sec:experimental_setup}

\vspace{-0.5em}
\paragraph{Datasets.}
We evaluate \name in two complementary settings: tool-integrated reasoning and long-horizon agentic tasks.
For tool-integrated reasoning, the training set contains 34,501 prompts drawn from 14,116 examples from the English subset of DAPO-Math, 10,385 examples from the code subset of Skywork-OR1-RL-Data, and 10,000 examples from MegaScience~\citep{abs-2503-14476,abs-2505-22312,abs-2507-16812}. 
We evaluate mathematical reasoning on AIME 2024--2026 and HMMT February 2026~\citep{aime24,aime25,aime26,abs-2605-00674}, scientific reasoning on GPQA~\citep{abs-2311-12022}, and code generation on LiveCodeBench v6~\citep{JainHGLYZWSSS25}.
For long-horizon interaction, we train and evaluate on the official splits of ALFWorld~\citep{ShridharYCBTH21} and ScienceWorld~\citep{WangJCA22}.
Both environments require the agent to condition successive actions on environment feedback, which makes them a direct test of robustness to compounding inter-turn error.

\vspace{-0.5em}
\paragraph{Baselines.}
We compare \name with the original student checkpoint (\textsc{Vanilla}), standard \textsc{OPD}~\citep{AgarwalVZSGGB24,Gu0WH24}, and five OPD methods that improve training stability through selective guidance or curricula: \textsc{SAGE-OPD}~\citep{abs-2606-19659}, \textsc{Guided-OPD}~\citep{abs-2606-15912}, \textsc{Relay-OPD}~\citep{abs-2607-26057}, \textsc{TCOD}~\citep{abs-2604-24005}, and \textsc{FutureBridge-OPD}~\citep{abs-2608-01953}.
All distillation methods share the same teacher and student initializations, prompts, rollout budget, and evaluation protocol, so the main comparison controls for differences in data and rollout configuration.
We additionally provide a comparison under a matched teacher-intervention budget in \Cref{app:teacher_computation_matched}.

\vspace{-0.5em}
\paragraph{Implementation Details.}
We use Qwen3-0.6B and Qwen3-1.7B as students and Qwen3-4B as the teacher~\citep{abs-2505-09388} and additional experiments with Qwen3-32B examine scaling to a stronger teacher.
Evaluation uses the student policy alone, with no teacher intervention.
All reported numbers are means over multiple independent runs (refer to the detailed experimental results), and the reported uncertainties are standard deviations across those runs. 
Complete hyperparameters and prompts are given in \Cref{app:training_parameters,app:prompts}. 

\begin{table}[t]
\centering 
\caption{Results on tool-integrated reasoning benchmarks. Bold and underlined values denote the best and second-best results, respectively.}
\label{tab:tir_results}
\resizebox{1.\linewidth}{!}{
\begin{tabular}{lcccccc}
\toprule
\multirow{2}{*}{\textbf{Method}} & \cellcolor{cyan!20} \textbf{AIME24} & \cellcolor{cyan!20} \textbf{AIME25} & \cellcolor{cyan!20} \textbf{AIME26} & \cellcolor{cyan!20} \textbf{HMMT26} & \cellcolor{Orchid!20} \textbf{GPQA} & \cellcolor{LimeGreen!20} \textbf{LiveCodeBench v6} \\
& \multicolumn{4}{c}{\cellcolor{cyan!20} \textbf{Avg@32}} & \multicolumn{1}{c}{\cellcolor{Orchid!20} \textbf{Avg@8}} & \multicolumn{1}{c}{\cellcolor{LimeGreen!20} \textbf{Avg@8}} \\
\midrule 
\multicolumn{7}{c}{\cellcolor{gray!10} \textit{\textbf{Qwen3-4B}}} \\
\textsc{Teacher} & $71.7_{\pm 0.3}$ & $65.0_{\pm 0.3}$ & $60.0_{\pm 0.3}$ & $39.8_{\pm 0.3}$ & $50.9_{\pm 0.3}$ & $70.5_{\pm 0.3}$ \\
\midrule
\multicolumn{7}{c}{\cellcolor{gray!10} \textit{\textbf{Qwen3-0.6B}}} \\
\textsc{Vanilla} & $8.3_{\pm 0.4}$ & $15.8_{\pm 0.6}$ & $9.6_{\pm 0.5}$ & $10.2_{\pm 0.4}$ & $26.0_{\pm 0.7}$ & $20.7_{\pm 0.5}$ \\
\textsc{OPD} & $11.2_{\pm 0.5}$ & $18.5_{\pm 0.5}$ & $12.4_{\pm 0.4}$ & $11.8_{\pm 0.6}$ & $27.1_{\pm 0.6}$ & $23.5_{\pm 0.7}$ \\
\textsc{SAGE-OPD} & $14.5_{\pm 0.6}$ & $21.0_{\pm 0.7}$ & $14.8_{\pm 0.6}$ & $13.9_{\pm 0.5}$ & $28.6_{\pm 0.5}$ & $26.2_{\pm 0.6}$ \\
\textsc{Guided-OPD} & $18.2_{\pm 0.7}$ & $25.1_{\pm 0.6}$ & $17.5_{\pm 0.7}$ & $16.2_{\pm 0.5}$ & $29.8_{\pm 0.6}$ & $31.4_{\pm 0.8}$ \\
\textsc{Relay-OPD} & $16.8_{\pm 0.5}$ & $23.4_{\pm 0.8}$ & $16.1_{\pm 0.5}$ & $14.7_{\pm 0.6}$ & $28.2_{\pm 0.7}$ & $29.1_{\pm 0.7}$ \\
\textsc{TCOD} & $20.4_{\pm 0.8}$ & $27.9_{\pm 0.7}$ & $19.3_{\pm 0.6}$ & $15.8_{\pm 0.7}$ & \underline{$31.2_{\pm 0.6}$} & $33.6_{\pm 0.8}$ \\
\textsc{FutureBridge-OPD} & \underline{$22.1_{\pm 0.7}$} & \underline{$30.2_{\pm 0.8}$} & \underline{$20.8_{\pm 0.7}$} & \underline{$17.5_{\pm 0.5}$} & $30.9_{\pm 0.5}$ & \underline{$35.8_{\pm 0.9}$} \\
\midrule
\name & \textbf{24.2}$_{\pm \mathbf{0.4}}$ & \textbf{33.5}$_{\pm \mathbf{0.4}}$ & \textbf{22.7}$_{\pm \mathbf{0.3}}$ & \textbf{18.6}$_{\pm \mathbf{0.3}}$ & \textbf{32.5}$_{\pm \mathbf{0.3}}$ & \textbf{38.8}$_{\pm \mathbf{0.4}}$ \\
\midrule
\multicolumn{7}{c}{\cellcolor{gray!10} \textit{\textbf{Qwen3-1.7B}}} \\
\textsc{Vanilla} & $42.9_{\pm 0.6}$ & $35.8_{\pm 0.7}$ & $32.8_{\pm 0.5}$ & $25.0_{\pm 0.6}$ & $34.2_{\pm 0.8}$ & $44.1_{\pm 0.7}$ \\
\textsc{OPD} & $44.5_{\pm 0.7}$ & $37.1_{\pm 0.6}$ & $33.9_{\pm 0.6}$ & $26.1_{\pm 0.5}$ & $35.4_{\pm 0.7}$ & $45.3_{\pm 0.6}$ \\
\textsc{SAGE-OPD} & $46.2_{\pm 0.8}$ & $38.6_{\pm 0.8}$ & $34.8_{\pm 0.7}$ & $27.0_{\pm 0.6}$ & $37.1_{\pm 0.6}$ & $46.7_{\pm 0.8}$ \\
\textsc{Guided-OPD} & $48.6_{\pm 0.7}$ & $41.0_{\pm 0.7}$ & $37.0_{\pm 0.8}$ & $28.9_{\pm 0.7}$ & $38.8_{\pm 0.7}$ & $48.4_{\pm 0.9}$ \\
\textsc{Relay-OPD} & $47.3_{\pm 0.6}$ & $39.8_{\pm 0.9}$ & $36.2_{\pm 0.6}$ & $27.8_{\pm 0.5}$ & $36.9_{\pm 0.8}$ & $47.5_{\pm 0.7}$ \\
\textsc{TCOD} & $49.8_{\pm 0.8}$ & \underline{$42.1_{\pm 0.8}$} & \underline{$38.3_{\pm 0.7}$} & $28.5_{\pm 0.6}$ & \underline{$40.2_{\pm 0.7}$} & $49.3_{\pm 0.8}$ \\
\textsc{FutureBridge-OPD} & \underline{$50.7_{\pm 0.7}$} & $41.4_{\pm 0.7}$ & $37.9_{\pm 0.6}$ & \underline{$29.1_{\pm 0.8}$} & $39.9_{\pm 0.6}$ & \underline{$49.9_{\pm 0.7}$} \\
\midrule
\name & \textbf{52.2}$_{\pm \mathbf{0.4}}$ & \textbf{44.7}$_{\pm \mathbf{0.5}}$ & \textbf{39.7}$_{\pm \mathbf{0.3}}$ & \textbf{31.4}$_{\pm \mathbf{0.6}}$ & \textbf{42.2}$_{\pm \mathbf{0.4}}$ & \textbf{52.8}$_{\pm \mathbf{0.4}}$ \\
\bottomrule
\end{tabular}
} 
\vspace{-1.0em}
\end{table}

\begin{table}[t]
\centering 
\caption{Success rates on long-horizon agentic benchmarks. We report results for six ALFWorld task categories and the average across ScienceWorld tasks. Bold and underlined values denote the best and second-best results, respectively.}
\label{tab:agentic_results}
\resizebox{1.\linewidth}{!}{
\begin{tabular}{lccccccc}
\toprule
\multirow{2}{*}{\textbf{Method}} & \multicolumn{6}{c}{\cellcolor{cyan!20} \textbf{ALFWorld}} & \cellcolor{LimeGreen!20} \\
& \cellcolor{cyan!20} \textbf{Pick} 
& \cellcolor{cyan!20} \textbf{Pick Two} 
& \cellcolor{cyan!20} \textbf{Pick \& Clean} 
& \cellcolor{cyan!20} \textbf{Pick \& Heat} 
& \cellcolor{cyan!20} \textbf{Pick \& Cool} 
& \cellcolor{cyan!20} \textbf{Look} 
& \cellcolor{LimeGreen!20} \multirow{-2}{*}{\textbf{ScienceWorld}} \\
\midrule 
\multicolumn{8}{c}{\cellcolor{gray!10} \textit{\textbf{Qwen3-4B}}} \\
\textsc{Teacher} & $87.5_{\pm 0.4}$ & $66.2_{\pm 0.4}$ & $35.5_{\pm 0.4}$ & $57.6_{\pm 0.4}$ & $46.4_{\pm 0.4}$ & $83.3_{\pm 0.4}$ & $26.1_{\pm 0.4}$ \\
\midrule
\multicolumn{8}{c}{\cellcolor{gray!10} \textit{\textbf{Qwen3-0.6B}}} \\
\textsc{Vanilla} & $4.2_{\pm 0.3}$ & $0.0$ & $1.1_{\pm 0.2}$ & $0.0$ & $0.0$ & $3.5_{\pm 0.3}$ & $0.8_{\pm 0.2}$ \\
\textsc{OPD} & $65.2_{\pm 0.4}$ & $18.5_{\pm 0.5}$ & $14.8_{\pm 0.4}$ & $4.5_{\pm 0.3}$ & $27.4_{\pm 0.5}$ & $6.8_{\pm 0.4}$ & $7.3_{\pm 0.5}$ \\
\textsc{SAGE-OPD} & $66.5_{\pm 0.5}$ & $23.2_{\pm 0.4}$ & $15.9_{\pm 0.3}$ & $9.8_{\pm 0.4}$ & $28.7_{\pm 0.4}$ & $13.5_{\pm 0.5}$ & $9.5_{\pm 0.4}$ \\
\textsc{Relay-OPD} & $67.8_{\pm 0.4}$ & $28.0_{\pm 0.5}$ & $17.2_{\pm 0.4}$ & $17.5_{\pm 0.5}$ & $30.5_{\pm 0.6}$ & $21.2_{\pm 0.4}$ & $10.2_{\pm 0.6}$ \\
\textsc{Guided-OPD} & $69.1_{\pm 0.5}$ & $33.6_{\pm 0.6}$ & $18.3_{\pm 0.5}$ & $24.6_{\pm 0.4}$ & $31.8_{\pm 0.5}$ & $30.8_{\pm 0.5}$ & $12.6_{\pm 0.5}$ \\
\textsc{TCOD} & $70.4_{\pm 0.4}$ & $39.8_{\pm 0.5}$ & $19.8_{\pm 0.4}$ & $32.7_{\pm 0.5}$ & \underline{$33.9_{\pm 0.4}$} & $39.5_{\pm 0.6}$ & $13.0_{\pm 0.4}$ \\
\textsc{FutureBridge-OPD} & \underline{$71.6_{\pm 0.5}$} & \underline{$44.5_{\pm 0.5}$} & \underline{$20.9_{\pm 0.4}$} & \underline{$37.8_{\pm 0.6}$} & $33.4_{\pm 0.5}$ & \underline{$46.8_{\pm 0.5}$} & \underline{$15.4_{\pm 0.5}$} \\
\midrule
\name & \textbf{73.5}$_{\pm \mathbf{0.4}}$ & \textbf{48.8}$_{\pm \mathbf{0.6}}$ & \textbf{22.4}$_{\pm \mathbf{0.5}}$ & \textbf{41.8}$_{\pm \mathbf{0.5}}$ & \textbf{36.2}$_{\pm \mathbf{0.6}}$ & \textbf{53.4}$_{\pm \mathbf{0.6}}$ & \textbf{18.7}$_{\pm \mathbf{0.6}}$ \\
\midrule
\multicolumn{8}{c}{\cellcolor{gray!10} \textit{\textbf{Qwen3-1.7B}}} \\
\textsc{Vanilla} & $62.5_{\pm 0.3}$ & $1.5_{\pm 0.2}$ & $12.9_{\pm 0.3}$ & $7.7_{\pm 0.2}$ & $3.6_{\pm 0.2}$ & $36.1_{\pm 0.4}$ & $6.3_{\pm 0.3}$ \\
\textsc{OPD} & $85.5_{\pm 0.5}$ & $38.7_{\pm 0.4}$ & $34.7_{\pm 0.5}$ & $21.7_{\pm 0.4}$ & $47.6_{\pm 0.5}$ & $22.2_{\pm 0.4}$ & $17.1_{\pm 0.5}$ \\
\textsc{SAGE-OPD} & $86.8_{\pm 0.4}$ & $43.5_{\pm 0.5}$ & $35.8_{\pm 0.4}$ & $27.4_{\pm 0.5}$ & $48.9_{\pm 0.6}$ & $29.8_{\pm 0.5}$ & $21.4_{\pm 0.4}$ \\
\textsc{Relay-OPD} & $88.1_{\pm 0.5}$ & $48.2_{\pm 0.6}$ & $37.1_{\pm 0.5}$ & $35.6_{\pm 0.4}$ & $50.8_{\pm 0.5}$ & $38.4_{\pm 0.6}$ & $17.8_{\pm 0.5}$ \\
\textsc{Guided-OPD} & $89.4_{\pm 0.4}$ & $53.9_{\pm 0.5}$ & $38.0_{\pm 0.4}$ & $42.8_{\pm 0.5}$ & $51.5_{\pm 0.4}$ & $48.6_{\pm 0.5}$ & $16.2_{\pm 0.6}$ \\
\textsc{TCOD} & $90.6_{\pm 0.5}$ & $60.1_{\pm 0.5}$ & $39.5_{\pm 0.5}$ & $51.3_{\pm 0.6}$ & \underline{$53.7_{\pm 0.5}$} & $58.2_{\pm 0.5}$ & $18.7_{\pm 0.5}$ \\
\textsc{FutureBridge-OPD} & \underline{$91.8_{\pm 0.4}$} & \underline{$64.7_{\pm 0.6}$} & \underline{$40.2_{\pm 0.3}$} & \underline{$57.5_{\pm 0.5}$} & $53.2_{\pm 0.4}$ & \underline{$66.5_{\pm 0.6}$} & \underline{$22.2_{\pm 0.4}$} \\
\midrule
\name & \textbf{93.7}$_{\pm \mathbf{0.5}}$ & \textbf{69.1}$_{\pm \mathbf{0.4}}$ & \textbf{41.2}$_{\pm \mathbf{0.3}}$ & \textbf{62.0}$_{\pm \mathbf{0.5}}$ & \textbf{56.0}$_{\pm \mathbf{0.4}}$ & \textbf{73.7}$_{\pm \mathbf{0.5}}$ & \textbf{27.5}$_{\pm \mathbf{0.5}}$ \\
\bottomrule
\end{tabular}
} 
\vspace{-1.0em}
\end{table}

\subsection{Main Results}
\label{sec:main_results}

\vspace{-0.5em}
\paragraph{\name consistently outperforms all distillation baselines.}
As shown in \Cref{tab:tir_results,tab:agentic_results}, \name attains the best mean score among the distillation methods in every reported cell, at both student sizes, spanning mathematical and scientific reasoning, code generation, and interactive decision making.
The advanced OPD baselines generally improve over standard \textsc{OPD}, yet \name remains ahead of the most competitive of them, \textsc{TCOD} and \textsc{FutureBridge-OPD}. 

\vspace{-0.5em}
\paragraph{\name improves tool-integrated reasoning across domains.}
On the tool-integrated benchmarks, \name improves over standard \textsc{OPD} and over the best-performing baseline on every dataset.
The gains extend beyond mathematics to GPQA and LiveCodeBench v6, which suggests that the method helps both structured reasoning and tool-supported generation.
The margin over \textsc{OPD} is larger for the 0.6B student than for the 1.7B student, consistent with adaptive teacher intervention being most useful when the student is less capable and therefore more likely to leave teacher-familiar contexts.

\vspace{-0.5em}
\paragraph{\name provides its largest gains on long-horizon agentic tasks.}
Long-horizon tasks like ALFWorld and ScienceWorld are representative tasks because actions change subsequent observations and errors propagate across turns. 
The largest gains therefore fall on the longest-horizon settings, ScienceWorld and the ALFWorld types that require the most environment interaction, which is the direction turn-adaptive intervention predicts under trajectory-level distribution shift. 
On some tasks, \name even outperforms its teacher, demonstrating the effectiveness of turn-adaptive intervention in mitigating student errors and enhancing long-horizon performance.

\subsection{Ablation Study}
\label{sec:ablation_study}

\begin{table}[t]
    \centering
    \caption{Ablation results on the long-horizon agentic benchmarks with the Qwen3-1.7B student. Each variant modifies one component of \name: replacing $k_3$ with $k_1$, removing importance weighting, or replacing reverse KL with forward KL.}
    \label{tab:ablation_study}
    \resizebox{\linewidth}{!}{%
    \begin{tabular}{lccccccc}
        \toprule
        \multirow{2}{*}{\textbf{Method}}
        & \multicolumn{6}{c}{\cellcolor{cyan!20}\textbf{ALFWorld}}
        & \cellcolor{LimeGreen!20} \\
        & \cellcolor{cyan!20}\textbf{Pick}
        & \cellcolor{cyan!20}\textbf{Pick Two}
        & \cellcolor{cyan!20}\textbf{Pick \& Clean}
        & \cellcolor{cyan!20}\textbf{Pick \& Heat}
        & \cellcolor{cyan!20}\textbf{Pick \& Cool}
        & \cellcolor{cyan!20}\textbf{Look}
        & \cellcolor{LimeGreen!20}\multirow{-2}{*}{\textbf{ScienceWorld}} \\
        \midrule
        \name
        & \textbf{93.7}$_{\pm \mathbf{0.5}}$ 
        & \textbf{69.1}$_{\pm \mathbf{0.4}}$ 
        & \textbf{41.2}$_{\pm \mathbf{0.3}}$ 
        & \textbf{62.0}$_{\pm \mathbf{0.5}}$ 
        & \textbf{56.0}$_{\pm \mathbf{0.4}}$ 
        & \textbf{73.7}$_{\pm \mathbf{0.5}}$ 
        & \textbf{27.5}$_{\pm \mathbf{0.5}}$ \\
        \hdashline
        \textit{w/} $k_1$ estimator
        & $89.5_{\pm 0.5}$ & $65.2_{\pm 0.4}$ & $39.4_{\pm 0.5}$ & $59.7_{\pm 0.5}$ & $51.3_{\pm 0.4}$ & $67.4_{\pm 0.5}$ & $23.8_{\pm 0.5}$ \\
        \textit{w/o} importance weighting
        & $86.2_{\pm 0.4}$ & $66.7_{\pm 0.4}$ & $36.8_{\pm 0.3}$ & $56.2_{\pm 0.4}$ & $46.4_{\pm 0.5}$ & $61.9_{\pm 0.4}$ & $21.5_{\pm 0.3}$ \\
        \textit{w/} forward KL
        & $91.3_{\pm 0.5}$ & $67.2_{\pm 0.4}$ & $38.2_{\pm 0.4}$ & $58.8_{\pm 0.5}$ & $54.9_{\pm 0.4}$ & $69.6_{\pm 0.5}$ & $23.1_{\pm 0.4}$ \\
        \bottomrule
    \end{tabular}%
    }
    \vspace{-1.0em}
\end{table}

\begin{figure}[t]
    \centering
    \includegraphics[width=1.\linewidth]{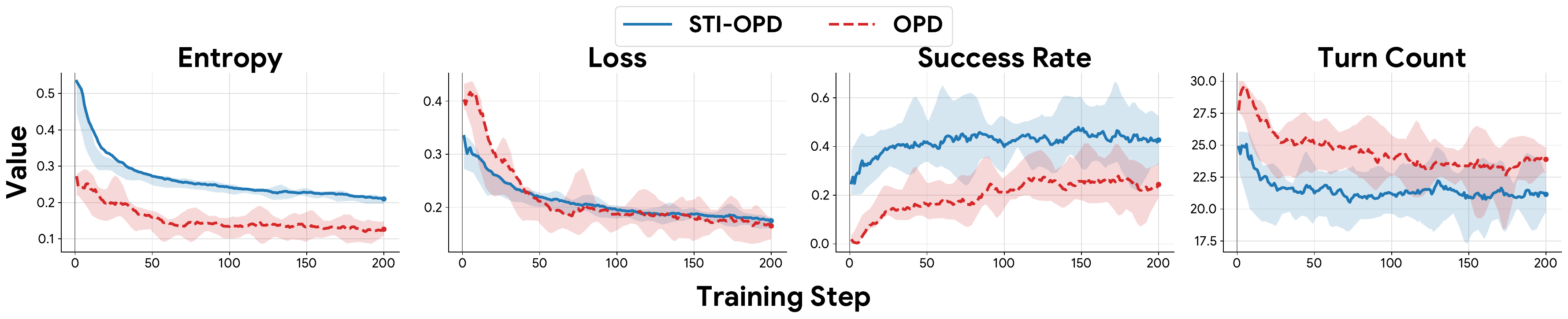}
    \caption{Training dynamics of \name and standard OPD on ALFWorld with Qwen3-1.7B as the student and Qwen3-4B as the teacher.}
    \label{fig:training_dynamics}
    \vspace{-1.0em}
\end{figure} 

\vspace{-0.5em}

\Cref{tab:ablation_study} isolates the contributions of the components in \name.
Replacing $k_3$ with $k_1$ reduces performance in both environments, supporting the use of a non-negative discrepancy estimator with lower variance near policy agreement for intervention.
We further compare the proposed stochastic intervention strategy against deterministic thresholding in \Cref{app:stochastic_vs_deterministic}. The stochastic strategy consistently performs better across ALFWorld and ScienceWorld without requiring a manually selected decision boundary.
Removing importance weighting causes the largest degradation on each reported task, highlighting the importance of correcting the sampling mismatch introduced by teacher-generated tokens.
Replacing reverse KL with forward KL also consistently reduces performance (we also study its impact on mitigating forgetting in \Cref{app:reverse_kl_forgetting}), favoring preservation of the conditional reverse-KL objective through importance weighting over changing the distillation objective.

\subsection{Mechanism, Scaling, and Cost Analysis of \name}
\label{sec:analysis}

\vspace{-0.5em}
\paragraph{Training Dynamics.}
As shown in \Cref{fig:training_dynamics}, \name achieves higher success rates early in training and maintains its advantage over standard OPD.
It also preserves higher policy entropy while requiring fewer interaction turns, suggesting greater policy diversity alongside more efficient task execution.
Although the training losses become similar at later stages, the success-rate gap persists, indicating that comparable distillation losses do not necessarily translate into comparable agentic performance.

\begin{figure}[t]
    \centering

    \begin{minipage}[t]{0.58\linewidth}
        \centering
        \includegraphics[width=\linewidth]{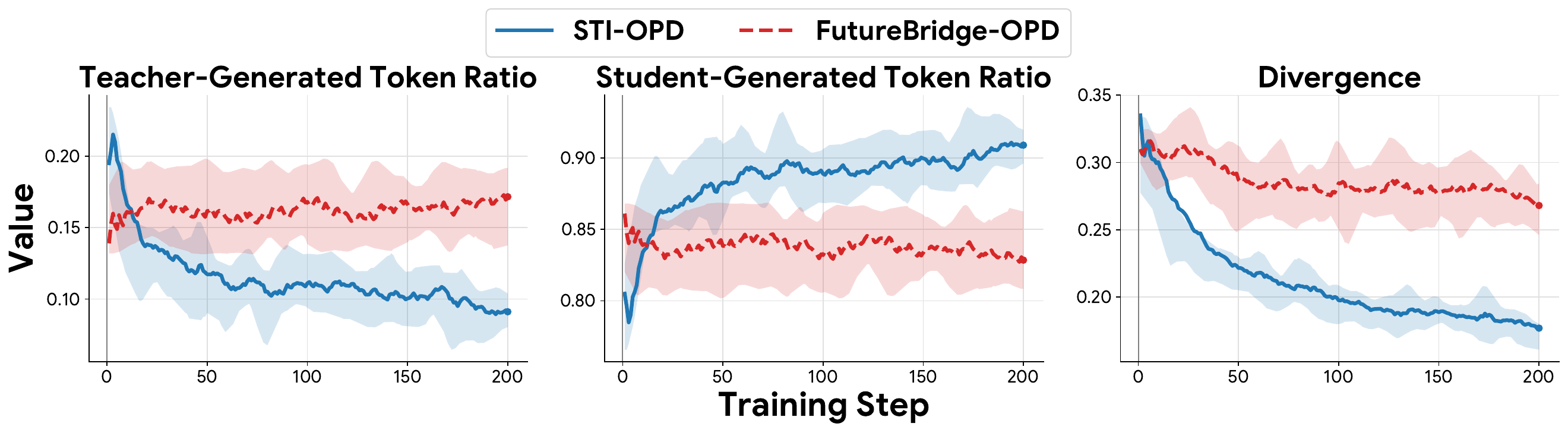}
        \caption{Intervention and policy discrepancy dynamics of \name and \textsc{FutureBridge-OPD} on ALFWorld, using Qwen3-1.7B as the student and Qwen3-4B as the teacher.}
        \label{fig:age_opd_dynamics}
    \end{minipage}
    \hfill
    \begin{minipage}[t]{0.4\linewidth}
        \centering
        \includegraphics[width=\linewidth]{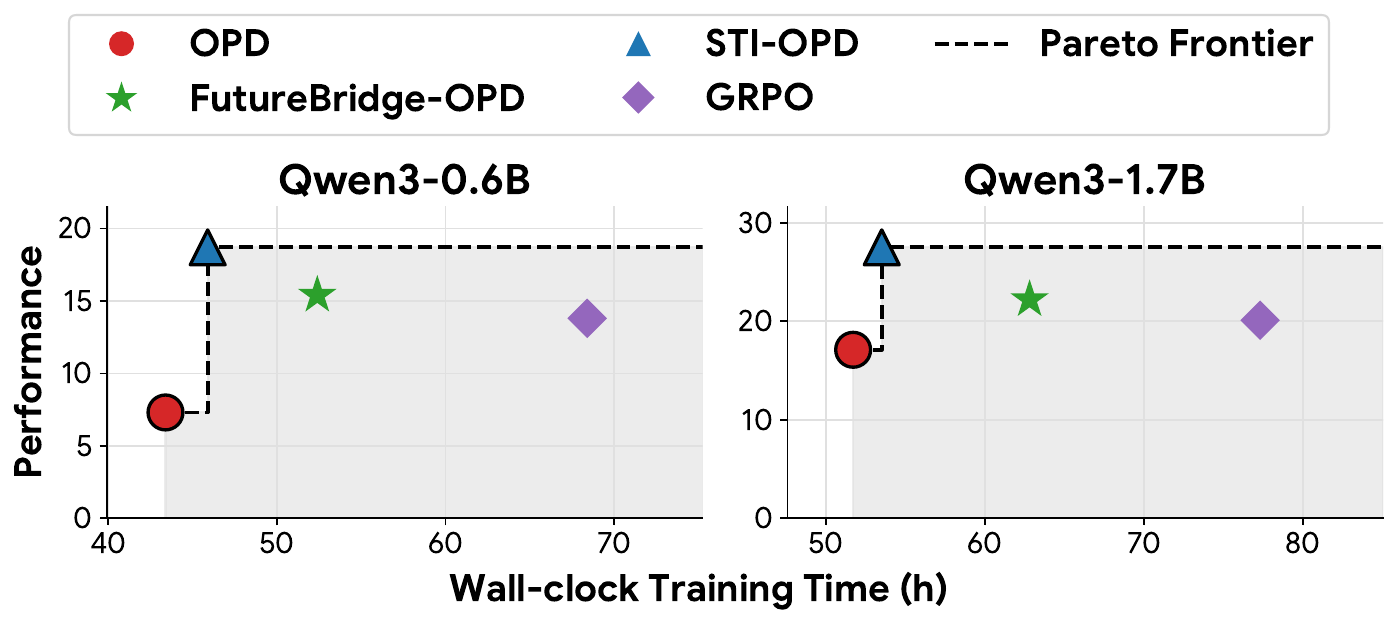}
        \caption{Performance versus training wall-clock time. The dashed line indicates the empirical Pareto frontier. }
        \label{fig:wallclock_time}
    \end{minipage}

    \vspace{-1.0em}
\end{figure}

\vspace{-0.5em}
\paragraph{Mechanism Analysis.}
\Cref{fig:age_opd_dynamics} shows that \name shifts progressively from teacher-generated to student-generated tokens as training proceeds, and that the shift is more pronounced than for \textsc{FutureBridge-OPD}.
Alongside this shift, the teacher--student policy discrepancy decreases toward zero, as measured by the turn-level score in \Cref{eq:normalized_turn_score}.
These coupled trends are what the intervention rule prescribes: the teacher takes over more often when policy discrepancy is large, then cedes control as the student aligns.
Training therefore approaches the student-controlled regime used at inference time on its own, without a predefined intervention schedule.
\Cref{app:case_studies} shows cases and discussions about the mechanism of \name.

\vspace{-0.5em}
\paragraph{Scalability.}
Replacing the Qwen3-4B teacher with large-scale Qwen3-32B lifts each method accordingly. 
\name keeps the lead on each ALFWorld subtask and on ScienceWorld: with the 1.7B student it reaches $97.8$ on \textsc{Look} and $56.3$ on ScienceWorld, against $90.4$ and $46.2$ for \textsc{FutureBridge-OPD} (see details in \Cref{app:larger_teacher_results}).
The mechanism is therefore not specific to the Qwen3-4B teacher, and its margin does not close as the teacher grows.

\vspace{-0.5em}
\paragraph{Cost Analysis.}
\Cref{fig:wallclock_time} shows that \name lies on the empirical Pareto frontier at both student sizes.
\name improves substantially over standard OPD with modest additional training time. 
Compared with \textsc{FutureBridge-OPD}, \name achieves higher performance at lower computational cost. 
To further rule out the possibility that these gains arise simply from greater teacher usage, we additionally compare intervention-based methods under the same teacher-intervention budget in \Cref{app:teacher_computation_matched}.
\name remains consistently better under this matched budget, indicating that its advantage comes from more effective allocation and utilization of teacher guidance rather than simply more frequent teacher intervention.
Its efficiency relative to \textsc{FutureBridge-OPD} is consistent with estimating discrepancy from the existing student proposal without additional lookahead rollouts. 
Moreover, the intervention frequency of \name decreases as the student policy aligns with the teacher, further reducing teacher dependence over training. 
Compared with \textsc{GRPO}, \name's advantage lies in its lower rollout computational cost, because it does not require group rollouts for reward normalization.
Together, these results support a favorable performance--cost trade-off.

\section{Conclusion}\label{sec:conclusion}

In this work, we studied on-policy distillation for multi-turn LLM agents, where an early student error changes the observations that follow and pushes the trajectory toward histories on which teacher supervision is less reliable. 
We introduced \name, which answers the two questions this setting raises: a per-sample non-negative $k_3$ estimate of teacher--student policy discrepancy sets a stochastic turn-level intervention probability, and an importance-sampling correction keeps supervision on teacher-generated responses faithful to the conditional reverse-KL objective.
Across tool-integrated reasoning and long-horizon agentic benchmarks, \name outperforms standard OPD and every intervention-based baseline we evaluate, at both student scales and with its largest margins on the longest-horizon tasks, while adding only modest wall-clock overhead over standard OPD.

\clearpage
\bibliography{refs}
\bibliographystyle{template}

\clearpage
\raggedbottom
\appendix 
\part{Appendix}
\startcontents
\printcontents{}{1}{\textbf{Contents of Appendix}\vskip3pt\hrule\vskip5pt}
\vskip3pt\hrule\vskip5pt
\clearpage

\section{Algorithmic Details of \name} 
\label{app:algorithmic_details}

We present the complete algorithmic details of \name in \Cref{alg:method}.

\begin{algorithm}[ht]
    \caption{\name: Stochastic Teacher Intervention for Agentic On-Policy Distillation}
    \label{alg:method}
    \small
    \begin{algorithmic}[1]
        \Require Task distribution $\mathcal{D}$; fixed teacher $\pi_T$;
        student $\pi_\theta$; maximum horizon $H$
        \Ensure Optimized student parameters $\theta$

        \While{not converged}
            \State Sample a task minibatch $\mathcal{Q}\sim\mathcal{D}$
            \State Initialize student and teacher buffers
            $\mathcal{B}_S,\mathcal{B}_T\gets\emptyset$

            \ForAll{$q\in\mathcal{Q}$}
                \State Sample $o_1\sim\rho_0(\cdot\mid q)$ and set
                $h_1\gets(q,o_1)$

                \For{$t=1,\ldots,H$}
                    \State Sample a student proposal
                    $\widetilde a_t=(\widetilde x_{t,1},\ldots,
                    \widetilde x_{t,\widetilde L_t})
                    \sim\pi_\theta(\cdot\mid h_t)$

                    \For{$k=1,\ldots,\widetilde L_t$}
                        \State $\widetilde c_{t,k}\gets
                        (h_t,\widetilde x_{t,<k})$
                        \State $r_{t,k}\gets
                        \dfrac{
                        \pi_T(\widetilde x_{t,k}\mid\widetilde c_{t,k})
                        }{
                        \pi_\theta(\widetilde x_{t,k}\mid\widetilde c_{t,k})
                        }$
                        \State $g_{t,k}\gets
                        r_{t,k}-1-\log r_{t,k}$
                    \EndFor

                    \State $s_t\gets
                    \widetilde L_t^{-1}
                    \sum_{k=1}^{\widetilde L_t}g_{t,k}$
                    \State $\alpha_t\gets s_t/(1+s_t)$
                    \State Sample
                    $z_t\sim\operatorname{Bernoulli}(\alpha_t)$

                    \If{$z_t=1$}
                        \State Sample a complete teacher turn
                        $a_t^T=(x^T_{t,1},\ldots,x^T_{t,L_t^T})
                        \sim\pi_T(\cdot\mid h_t)$

                        \For{$k=1,\ldots,L_t^T$}
                            \State $c^T_{t,k}\gets(h_t,x^T_{t,<k})$
                            \State Add $(x^T_{t,k},c^T_{t,k})$
                            to $\mathcal{B}_T$
                        \EndFor

                        \State Execute $a_t\gets a_t^T$
                    \Else
                        \For{$k=1,\ldots,\widetilde L_t$}
                            \State Add
                            $(\widetilde x_{t,k},\widetilde c_{t,k})$
                            to $\mathcal{B}_S$
                        \EndFor

                        \State Execute $a_t\gets\widetilde a_t$
                    \EndIf

                    \State Observe
                    $o_{t+1}\sim P(\cdot\mid h_t,a_t)$
                    and update $h_{t+1}$

                    \If{the environment terminates}
                        \State \textbf{break}
                    \EndIf
                \EndFor
            \EndFor

            \State Define
            $\ell_\theta(x,c)\gets
            \log\pi_\theta(x\mid c)-\log\pi_T(x\mid c)$

            \State $\mathcal{L}_S\gets
            \displaystyle
            \sum_{(x,c)\in\mathcal{B}_S}
            \ell_\theta(x,c)$

            \State $\mathcal{L}_T\gets
            \displaystyle
            \sum_{(x,c)\in\mathcal{B}_T}
            \frac{\pi_\theta(x\mid c)}
                 {\pi_T(x\mid c)}
            \ell_\theta(x,c)$

            \State $\widehat{\mathcal{L}}_{\name}\gets
            \dfrac{\mathcal{L}_S+\mathcal{L}_T}
            {|\mathcal{B}_S|+|\mathcal{B}_T|}$

            \State Update $\theta$ using
            $\nabla_\theta\widehat{\mathcal{L}}_{\name}$
        \EndWhile
    \end{algorithmic}
\end{algorithm}

\section{Additional Experimental Details}
\label{app:experimental_details}

\subsection{Datasets}
\label{app:datasets}

\paragraph{Training Corpus.}
Our tool-integrated reasoning corpus combines mathematical, coding, and scientific problems from three public sources.  
We use the English subset of DAPO-Math, the code subset of Skywork-OR1-RL-Data, and 10,000 examples from MegaScience. 
The resulting mixture contains 34,501 training prompts: 14,116 from DAPO-Math, 10,385 from Skywork-OR1-RL-Data, and 10,000 from MegaScience.  
The original responses associated with these datasets are not used as fixed distillation targets, instead, each prompt initiates a fresh rollout under the policy specified by the corresponding distillation method.

\paragraph{DAPO-Math.}
DAPO-Math-17K is a collection of 17,000 mathematical reasoning prompts curated from Web sources and official competition materials~\citep{abs-2503-14476}. 
The source dataset transforms each problem so that its target is an integer-valued answer, which enables reliable rule-based verification while retaining a broad range of competition-style reasoning patterns. 
We retain only its English-language examples for training.

\paragraph{Skywork-OR1-RL-Data.}
Skywork-OR1-RL-Data was developed for reinforcement-learning post-training of long-chain-of-thought reasoning models~\citep{abs-2505-22312}.  
Its released code portion contains 13,700 executable programming problems after quality filtering and deduplication, including 2,700 problems derived from LeetCode and 11,000 from TACO\@. 
We use this code subset to expose the student to algorithm design, program synthesis, and execution-based correctness signals.

\paragraph{MegaScience.}
MegaScience is a 1.25-million-example mixture of open scientific reasoning datasets spanning seven disciplines~\citep{abs-2507-16812}. 
It is designed to broaden post-training beyond mathematics and coding by providing diverse, verifiable scientific questions. 
We sample 10,000 examples from MegaScience and include them as the scientific component of our training mixture.

\paragraph{AIME 2024--2026.}
The American Invitational Mathematics Examination (AIME) consists of 15 short-answer problems whose answers are integers between 000 and 999. 
We use the 2024, 2025, and 2026 contests~\citep{aime24,aime25,aime26}. 
Their recency and emphasis on multi-step algebraic, geometric, combinatorial, and number-theoretic reasoning make them useful for evaluating mathematical generalization across successive competition years.

\paragraph{HMMT February 2026.}
The February 2026 Harvard--MIT Mathematics Tournament benchmark contains recent olympiad-style problems covering several areas of high-school mathematics. 
We include it as an additional out-of-corpus evaluation of challenging mathematical reasoning~\citep{abs-2605-00674}.

\paragraph{GPQA.}
GPQA is a graduate-level multiple-choice benchmark comprising 448 questions written and validated by domain experts in biology, physics, and chemistry~\citep{abs-2311-12022}. 
The questions are designed to remain difficult even for skilled non-experts with unrestricted Web access. 
GPQA therefore evaluates whether the distilled student transfers scientific reasoning ability beyond the scientific prompts seen during post-training.

\paragraph{LiveCodeBench v6.}
LiveCodeBench is a continuously updated code benchmark constructed from recent problems collected from LeetCode, AtCoder, and Codeforces~\citep{JainHGLYZWSSS25}. 
It evaluates generated programs with executable test cases and reduces contamination risk by introducing newly released contest problems over time. 
We use version~6 for all code-generation evaluations.

\paragraph{ALFWorld.}
ALFWorld aligns a text-based simulator with embodied household tasks from the ALFRED benchmark~\citep{ShridharYCBTH21}. 
An agent receives a natural-language goal and must interact with the environment through a sequence of textual actions, such as navigation, object manipulation, cleaning, heating, cooling, and placement.  Because every action changes the observations available at later turns, early mistakes can alter the remainder of the trajectory.

\paragraph{ScienceWorld.}
ScienceWorld is an interactive text environment grounded in an elementary science curriculum~\citep{WangJCA22}. 
Its tasks require agents to explore the environment, manipulate objects, conduct simple experiments, and use the resulting observations to achieve a scientific goal. 
The environment tests both procedural knowledge and long-horizon adaptation to feedback, making it a
complementary setting to ALFWorld for studying inter-turn error accumulation.

Unless stated otherwise, all methods use identical training instances and evaluation tasks within each setting. 
This controls for variation in task composition and isolates the effect of the distillation procedure. 

\begin{table*}[t]
    \centering
    \caption{Training and rollout parameters.  Unless stated otherwise, these
    settings are shared by \name and all distillation baselines.}
    \label{tab:training_parameters}
    \small
    \resizebox{\linewidth}{!}{
    \begin{tabular}{@{}lll@{}}
        \toprule
        \textbf{Category} & \textbf{Parameter} & \textbf{Value} \\
        \midrule
        Model
        & Student
        & Qwen3-0.6B; Qwen3-1.7B \\
        & Teacher
        & Qwen3-4B; Qwen3-32B \\
        \midrule
        Data 
        & \multicolumn{2}{c}{\textit{Tool-Integrated Reasoning}} \\
        \cmidrule(lr){2-3}
        & \# Training prompts
        & 34,501 \\
        & Training epochs
        & 1 \\
        & Batch size
        & 128 \\ 
        \cmidrule(lr){2-3}
        & \multicolumn{2}{c}{\textit{ALFWorld}} \\
        \cmidrule(lr){2-3}
        & \# Training prompts
        & 3,553 \\
        & Training steps
        & 200 \\
        & Batch size
        & 32 \\
        \cmidrule(lr){2-3}
        & \multicolumn{2}{c}{\textit{ScienceWorld}} \\
        \cmidrule(lr){2-3}
        & \# Training prompts
        & 2,294 \\
        & Training steps
        & 200 \\
        & Batch size
        & 32 \\
        \midrule
        Rollout
        & Rollouts per training prompt
        & 1 \\
        & Sampling temperature
        & 1.0 \\
        & Top-$p$
        & 1.0 \\
        & Maximum response length per turn
        & 32,768 tokens \\
        & Maximum model context length
        & 40,960 tokens \\
        \cmidrule(lr){2-3}
        & \multicolumn{2}{c}{\textit{Maximum Turns}} \\
        \cmidrule(lr){2-3}
        & Tool-integrated reasoning
        & 5 \\
        & ALFWorld
        & 30 \\
        & ScienceWorld
        & 30 \\
        \midrule
        Optimization
        & Learning rate
        & $1\times10^{-6}$ \\
        & Gradient clipping
        & 1.0 \\
        \midrule
        Evaluation
        & Policy used at inference
        & Student only; no teacher intervention \\
        & Reported aggregation
        & Mean across different runs \\
        \midrule
        Hardware
        & GPU 
        & 4 NVIDIA GB200 GPUs \\
        & CPU
        & Neoverse-v2 \\
        \midrule
        Software
        & CUDA
        & 13.0 \\
        & Pytorch~\citep{PaszkeGMLBCKLGA19}
        & 2.11.0 \\
        & vLLM~\citep{KwonLZ0ZY0ZS23} 
        & 0.23.0 \\
        & Flash-Attention~\citep{DaoFERR22} 
        & 0.8.3 \\
        & Trinity-RFT~\citep{abs-2505-17826} 
        & 0.6.0 \\
        \bottomrule
    \end{tabular}
    }
\end{table*} 

\subsection{Baselines}
\label{app:baselines}

\paragraph{\textsc{Vanilla}.}
\textsc{Vanilla} is the original student checkpoint without additional post-training. 
It measures the capability supplied by the pretrained model and provides a reference for quantifying the net effect of distillation.

\paragraph{\textsc{OPD}.}
\textsc{OPD} denotes standard on-policy distillation~\citep{AgarwalVZSGGB24,Gu0WH24}. 
It collects rollouts from the student and minimizes the token-level reverse KL divergence to a fixed teacher on the student-visited contexts. 
It therefore reduces the sequence-level mismatch between offline training and student inference, but does not explicitly control trajectory drift in multi-turn interaction.

\paragraph{\textsc{SAGE-OPD}.}
\textsc{SAGE-OPD} selectively allocates teacher supervision at the turn level~\citep{abs-2606-19659}.  After observing environment feedback, it uses teacher judgment to decide whether a student turn should be skipped or supervised. 
It further weights token-level distillation by teacher confidence and normalizes the loss to preserve the scale of the standard OPD objective.

\paragraph{\textsc{Guided-OPD}.}
\textsc{Guided-OPD} mixes teacher- and student-generated turns within each trajectory~\citep{abs-2606-15912}. 
A curriculum begins with frequent teacher intervention to keep early rollouts close to the teacher-induced history distribution and gradually decays the intervention probability to zero, recovering student-only rollouts toward the end of training.

\paragraph{\textsc{Relay-OPD}.}
\textsc{Relay-OPD} targets prefix failures by detecting locations where teacher and student continuations exhibit different tendencies~\citep{abs-2607-26057}. 
At a detected trigger, the teacher temporarily takes over to generate a short teacher relay segment, after which control returns to the student. 
A limited relay budget concentrates guidance at selected positions while restricting the deviation from student-generated trajectories.

\paragraph{\textsc{TCOD}.}
\textsc{TCOD} introduces a temporal curriculum over interaction depth~\citep{abs-2604-24005}. 
It initially exposes the student to short trajectory prefixes and progressively increases the number of student-controlled turns. 
This curriculum is designed to prevent early inter-turn errors from immediately pushing training trajectories far from the teacher-induced distribution.

\paragraph{\textsc{FutureBridge-OPD}.}
\textsc{FutureBridge-OPD} evaluates teacher guidance through its downstream effect on the student trajectory~\citep{abs-2608-01953}. 
At a high-discrepancy state, it inserts a short teacher-generated bridge and then examines the subsequent student continuation to determine whether the bridge increases the density of useful distillation signals.
It thereby selects interventions using future trajectory validation rather than only the current turn's discrepancy.

All methods use the same student initialization, teacher checkpoint, training prompts, rollout budget, and evaluation protocol for a given comparison.  
We also apply the same context and response limits across methods. 
This setup ensures that performance differences are attributable to the distillation and intervention strategies rather than to differences in models or data budgets. 

\subsection{Training Configuration and Parameters}
\label{app:training_parameters}

We use Qwen3-0.6B and Qwen3-1.7B as students and Qwen3-4B as the default teacher~\citep{abs-2505-09388}.  
We additionally use Qwen3-32B as a stronger teacher to study whether \name scales with teacher capability. 
Shared training and rollout parameters are summarized in \Cref{tab:training_parameters}.

For every training prompt, a distillation method collects one rollout using temperature~1.0 and top-$p$~1.0. 
At each turn, the model generates at most 32,768 response tokens and maximum model context length is 40,960 tokens.  
We cap tool-integrated reasoning trajectories at five turns and trajectories in ALFWorld and ScienceWorld at 30 turns. 
All methods are trained for one epoch for TIR and 200 steps for ALFWorld and ScienceWorld with a learning rate of $1\times10^{-6}$.

During \name training, the teacher scores each student-proposed turn, the intervention decision is sampled according to \Cref{eq:intervention_probability}, and the student is optimized on teacher-generated tokens using the importance-weighted objective in \Cref{eq:intervention_corrected_objective}. 
The teacher is used only during training and evaluation runs the student policy alone. 
We report the mean over multiple independent runs, with standard deviations taken across those runs, and
reproduce all task prompts verbatim in \Cref{app:prompts}. 

In terms of engineering implementation, we optimize \name based on a policy gradient-style workflow. 
Specifically, we treat $-\ell_\theta$ in \Cref{eq:intervention_corrected_objective} as the reward and then compute the PPO-style loss. 
This allows us to leverage well-established techniques such as clipping to achieve more stable optimization. 
For instance, $\omega^{T}_{t,k}$ in \Cref{eq:intervention_corrected_objective} will be clipped to enable stable and controllable step-wise optimization. 

\subsection{Prompt Templates}
\label{app:prompts}

For reproducibility, we provide the complete task prompts used in our experiments. 
Mathematical and scientific reasoning examples share the same \texttt{MATH\_PROMPT}. 
Only the task-specific \texttt{\{question\}} field is substituted. 
Code-generation examples use a separate template that enforces a Python-only output format. 
ALFWorld and ScienceWorld use environment-specific system prompts that define the interaction protocol and valid action space.

For interactive tasks, each environment observation is supplied in a user message prefixed by \texttt{Observation:}. In ALFWorld, an observation containing \texttt{Nothing happens.} is additionally followed by a reminder to check action validity and output formatting. 
In ScienceWorld, the initial observation is also prefixed by the task description. 
The prompt text below is otherwise reproduced verbatim from the training workflow.

\begin{prompt}{Mathematical and Scientific Reasoning}{math-science}
{\footnotesize\ttfamily\raggedright
\agePromptLine{Analyze and solve the following mathematics problem step by step.}
\agePromptBlank
\agePromptLine{Question:}
\agePromptLine{{question}}
\agePromptBlank
\agePromptLine{You may use available tools to perform precise calculations, validate intermediate results, and verify your final answer when helpful.}
\agePromptBlank
\agePromptLine{Instructions:}
\agePromptLine{1. First, analyze the problem and formulate a high-level solution strategy.}
\agePromptLine{2. Carry out the solution step by step, using tools only when they improve accuracy or efficiency.}
\agePromptLine{3. Before finalizing, verify that the result satisfies all conditions in the problem.}
\agePromptLine{4. Once you have determined the final answer, do not call any additional tools.}
\agePromptBlank
\agePromptLine{Your final response must contain the answer in exactly this format:}
\noindent\textbackslash boxed\{\{The final answer goes here.\}\}\par
}
\end{prompt}

\begin{prompt}{Code Generation}{code-generation}
{\footnotesize\ttfamily\raggedright
\agePromptLine{You will be given a programming problem specification. Write a correct Python program that solves the problem and passes all test cases.}
\agePromptBlank
\agePromptLine{Question:}
\agePromptLine{{question}}
\agePromptBlank
\agePromptLine{Instructions:}
\agePromptLine{1. Analyze the problem and formulate a high-level solution strategy.}
\agePromptLine{2. Derive the algorithm, considering correctness, edge cases, and complexity.}
\agePromptLine{3. You may use available tools to validate reasoning or test edge cases when helpful.}
\agePromptLine{4. Do not execute the program directly on the provided sample inputs.}
\agePromptLine{5. Once you are confident that the solution is correct, do not call additional tools.}
\agePromptBlank
\agePromptLine{The program must read from stdin, write only the required output to stdout, be self-contained, and be compatible with Python.}
\agePromptBlank
\agePromptLine{Return only the final code in this format:}
\noindent\textasciigrave\textasciigrave\textasciigrave python\par
\noindent\# YOUR CODE HERE\par
\noindent\textasciigrave\textasciigrave\textasciigrave\par
}
\end{prompt}

\begin{prompt}{ALFWorld System}{alfworld-system}
{\footnotesize\ttfamily\raggedright
\agePromptLine{You are an agent interacting with a virtual text-based environment.}
\agePromptBlank
\noindent\#\# Notes:\par
\agePromptLine{At each step, you should first think then perform action to fulfill the instruction. You should ALWAYS wrap your thinking with the <think> </think> tag and wrap your action with the <action> </action> tag.}
\agePromptLine{You should ALWAYS take one action each step.}
\agePromptLine{DO NOT try to interact with the user at anytime. Finish the task by yourself.}
\agePromptBlank
\noindent\#\# Action Format:\par
\agePromptLine{Below are the available commands you can use:}
\agePromptLine{  look:                             look around your current location}
\agePromptLine{  inventory:                        check your current inventory(you can only have 1 item in your inventory)}
\agePromptLine{  go to (receptacle):               move to a receptacle}
\agePromptLine{  open (receptacle):                open a receptacle}
\agePromptLine{  close (receptacle):               close a receptacle}
\agePromptLine{  take (object) from (receptacle):  take an object from a receptacle}
\agePromptLine{  move (object) to (receptacle):    place an object in or on a receptacle}
\agePromptLine{  examine (something):              examine a receptacle or an object}
\agePromptLine{  use (object):                     use an object}
\agePromptLine{  heat (object) with (receptacle):  heat an object using a receptacle}
\agePromptLine{  clean (object) with (receptacle): clean an object using a receptacle}
\agePromptLine{  cool (object) with (receptacle):  cool an object using a receptacle}
\agePromptLine{  slice (object) with (object):     slice an object using a sharp object}
\agePromptBlank
\agePromptLine{For example your output should be like this:}
\agePromptLine{<think> To solve the task, I need first to ... </think><action>go to cabinet 1</action>}
}
\end{prompt}

\begin{prompt}{ScienceWorld System}{scienceworld-system}
{\footnotesize\ttfamily\raggedright
\agePromptLine{You are an agent, your job is to do some scientific experiment in a virtual text-based environment.}
\agePromptBlank
\noindent\#\# Notes:\par
\agePromptLine{At each step, you should first think then perform action to fulfill the instruction. You should ALWAYS wrap your thinking with the <think> </think> tag and wrap your action with the <action> </action> tag.}
\agePromptLine{You should ALWAYS take one action each step.}
\agePromptLine{DO NOT try to interact with the user at anytime. Finish the task by yourself.}
\agePromptBlank
\noindent\#\# Action Format:\par
\agePromptLine{Below are the available commands you can use:}
\agePromptLine{    open OBJ: open a container}
\agePromptLine{    close OBJ: close a container}
\agePromptLine{    activate OBJ: activate a device}
\agePromptLine{    deactivate OBJ: deactivate a device}
\agePromptLine{    connect OBJ to OBJ: connect electrical components}
\agePromptLine{    disconnect OBJ: disconnect electrical components}
\agePromptLine{    use OBJ [on OBJ]: use a device/item}
\agePromptLine{    look around: describe the current room}
\agePromptLine{    examine OBJ: describe an object in detail}
\agePromptLine{    look at OBJ: describe a container's contents}
\agePromptLine{    read OBJ: read a note or book}
\agePromptLine{    move OBJ to OBJ: move an object to a container}
\agePromptLine{    pick up OBJ: move an object to the inventory}
\agePromptLine{    pour OBJ into OBJ: pour a liquid into a container}
\agePromptLine{    mix OBJ: chemically mix a container}
\agePromptLine{    teleport to LOC: teleport to a specific room}
\agePromptLine{    focus on OBJ: signal intent on a task object}
\agePromptLine{    wait: task no action for 10 steps}
\agePromptLine{    wait1: task no action for a step}
\agePromptBlank
\agePromptLine{For example your output should be like this:}
\agePromptLine{<think> Now I will check the bedroom ... </think><action>teleport to bedroom</action>}
}
\end{prompt}

\section{Additional Experimental Results}
\label{app:additional_results}

\subsection{\name Results with Larger Teacher} 
\label{app:larger_teacher_results} 

To demonstrate the effectiveness of \name with a larger teacher model, we conduct experiments on ALFWorld and ScienceWorld using Qwen3-32B as the teacher. 
We focus on these benchmarks due to computational constraints and our primary interest in long-horizon agentic tasks.

\paragraph{\name Remains Effective With A Larger Teacher.} 
As shown in \Cref{tab:larger_teacher_results}, \name achieves the highest success rate among the
distillation methods in every reported ALFWorld category and on ScienceWorld, at both student
sizes, when Qwen3-32B is used as the teacher. Its advantage holds not only over standard OPD but also over the stronger intervention-based baselines, indicating that the benefit of discrepancy-adaptive intervention is preserved as teacher capacity increases. The framework is therefore not specific to the Qwen3-4B teacher used in the main experiments. 

\paragraph{The Largest Gains Occur on Tasks with Longer Dependencies.}
Relative to the best competing method, the margins are most pronounced on ScienceWorld and
\textsc{Look} ($+10.1$ and $+7.4$ points for the 1.7B student), while remaining positive on every
task category. These settings require information and action outcomes to be propagated across multiple turns, making them especially sensitive to early trajectory deviations. This pattern is consistent with the intended role of \name: stronger teacher supervision is most useful when combined with adaptive intervention at histories where student errors would otherwise affect subsequent decisions. 

\begin{table}[t]
\centering 
\caption{Success rates on long-horizon agentic benchmarks with Qwen3-32B as teacher. We report results for six ALFWorld task categories and the average across ScienceWorld tasks. Bold and underlined values denote the best and second-best results, respectively.}
\label{tab:larger_teacher_results}
\resizebox{1.\linewidth}{!}{
\begin{tabular}{lccccccc}
\toprule
\multirow{2}{*}{\textbf{Method}} & \multicolumn{6}{c}{\cellcolor{cyan!20} \textbf{ALFWorld}} & \cellcolor{LimeGreen!20} \\
& \cellcolor{cyan!20} \textbf{Pick} 
& \cellcolor{cyan!20} \textbf{Pick Two} 
& \cellcolor{cyan!20} \textbf{Pick \& Clean} 
& \cellcolor{cyan!20} \textbf{Pick \& Heat} 
& \cellcolor{cyan!20} \textbf{Pick \& Cool} 
& \cellcolor{cyan!20} \textbf{Look} 
& \cellcolor{LimeGreen!20} \multirow{-2}{*}{\textbf{ScienceWorld}} \\
\midrule 
\multicolumn{8}{c}{\cellcolor{gray!10} \textit{\textbf{Qwen3-32B}}} \\
\textsc{Teacher} & $91.7_{\pm 0.4}$ & $72.1_{\pm 0.3}$ & $62.1_{\pm 0.3}$ & $80.4_{\pm 0.5}$ & $65.5_{\pm 0.4}$ & $93.1_{\pm 0.3}$ & $47.8_{\pm 0.4}$ \\
\midrule
\multicolumn{8}{c}{\cellcolor{gray!10} \textit{\textbf{Qwen3-0.6B}}} \\
\textsc{Vanilla} & $4.2_{\pm 0.3}$ & $0.0_{\pm 0.2}$ & $1.1_{\pm 0.2}$ & $0.0_{\pm 0.2}$ & $0.0_{\pm 0.2}$ & $3.5_{\pm 0.3}$ & $0.8_{\pm 0.2}$ \\
\textsc{OPD} & $66.9_{\pm 0.4}$ & $22.4_{\pm 0.5}$ & $34.7_{\pm 0.4}$ & $13.3_{\pm 0.3}$ & $42.2_{\pm 0.5}$ & $16.0_{\pm 0.4}$ & $22.6_{\pm 0.5}$ \\
\textsc{SAGE-OPD} & $68.3_{\pm 0.5}$ & $27.5_{\pm 0.4}$ & $37.9_{\pm 0.3}$ & $21.5_{\pm 0.4}$ & $44.6_{\pm 0.4}$ & $25.7_{\pm 0.5}$ & $25.0_{\pm 0.4}$ \\
\textsc{Relay-OPD} & $69.8_{\pm 0.4}$ & $32.9_{\pm 0.5}$ & $41.0_{\pm 0.4}$ & $31.3_{\pm 0.5}$ & $47.2_{\pm 0.6}$ & $35.6_{\pm 0.4}$ & $25.9_{\pm 0.6}$ \\
\textsc{Guided-OPD} & $71.2_{\pm 0.5}$ & $38.9_{\pm 0.6}$ & $44.0_{\pm 0.5}$ & $40.0_{\pm 0.4}$ & $49.9_{\pm 0.5}$ & $47.5_{\pm 0.5}$ & $29.2_{\pm 0.5}$ \\
\textsc{TCOD} & $72.6_{\pm 0.4}$ & $45.7_{\pm 0.5}$ & $47.1_{\pm 0.4}$ & $49.8_{\pm 0.5}$ & \underline{$53.1_{\pm 0.4}$} & $57.5_{\pm 0.6}$ & $29.9_{\pm 0.4}$ \\
\textsc{FutureBridge-OPD} & \underline{$73.8_{\pm 0.5}$} & \underline{$51.0_{\pm 0.5}$} & \underline{$49.2_{\pm 0.4}$} & \underline{$56.5_{\pm 0.6}$} & $52.5_{\pm 0.5}$ & \underline{$65.7_{\pm 0.5}$} & \underline{$32.4_{\pm 0.5}$} \\
\midrule
\name & \textbf{75.6}$_{\pm \mathbf{0.4}}$ & \textbf{55.6}$_{\pm \mathbf{0.6}}$ & \textbf{52.2}$_{\pm \mathbf{0.5}}$ & \textbf{61.3}$_{\pm \mathbf{0.5}}$ & \textbf{56.4}$_{\pm \mathbf{0.6}}$ & \textbf{72.8}$_{\pm \mathbf{0.6}}$ & \textbf{40.5}$_{\pm \mathbf{0.6}}$ \\
\midrule
\multicolumn{8}{c}{\cellcolor{gray!10} \textit{\textbf{Qwen3-1.7B}}} \\
\textsc{Vanilla} & $62.5_{\pm 0.3}$ & $1.5_{\pm 0.2}$ & $12.9_{\pm 0.3}$ & $7.7_{\pm 0.2}$ & $3.6_{\pm 0.2}$ & $36.1_{\pm 0.4}$ & $6.3_{\pm 0.3}$ \\
\textsc{OPD} & $87.2_{\pm 0.4}$ & $42.6_{\pm 0.5}$ & $54.6_{\pm 0.4}$ & $30.5_{\pm 0.5}$ & $62.4_{\pm 0.5}$ & $31.4_{\pm 0.4}$ & $32.4_{\pm 0.5}$ \\
\textsc{SAGE-OPD} & $88.6_{\pm 0.5}$ & $47.8_{\pm 0.4}$ & $57.8_{\pm 0.5}$ & $39.1_{\pm 0.4}$ & $64.8_{\pm 0.6}$ & $42.0_{\pm 0.5}$ & $36.9_{\pm 0.4}$ \\
\textsc{Relay-OPD} & $90.1_{\pm 0.4}$ & $53.1_{\pm 0.6}$ & $60.9_{\pm 0.3}$ & $49.4_{\pm 0.5}$ & $67.5_{\pm 0.4}$ & $52.8_{\pm 0.6}$ & $33.5_{\pm 0.5}$ \\
\textsc{Guided-OPD} & $91.5_{\pm 0.5}$ & $59.2_{\pm 0.5}$ & $63.7_{\pm 0.4}$ & $58.2_{\pm 0.4}$ & $69.6_{\pm 0.5}$ & $65.3_{\pm 0.4}$ & $42.8_{\pm 0.6}$ \\
\textsc{TCOD} & $92.8_{\pm 0.4}$ & $66.0_{\pm 0.5}$ & $66.8_{\pm 0.5}$ & $68.4_{\pm 0.6}$ & \underline{$72.9_{\pm 0.5}$} & $76.2_{\pm 0.5}$ & $45.6_{\pm 0.5}$ \\
\textsc{FutureBridge-OPD} & \underline{$94.0_{\pm 0.5}$} & \underline{$71.2_{\pm 0.6}$} & \underline{$68.5_{\pm 0.4}$} & \underline{$76.2_{\pm 0.5}$} & $72.3_{\pm 0.4}$ & \underline{$90.4_{\pm 0.6}$} & \underline{$46.2_{\pm 0.4}$} \\
\midrule
\name & \textbf{95.8}$_{\pm \mathbf{0.5}}$ & \textbf{75.9}$_{\pm \mathbf{0.5}}$ & \textbf{71.0}$_{\pm \mathbf{0.3}}$ & \textbf{81.5}$_{\pm \mathbf{0.6}}$ & \textbf{76.2}$_{\pm \mathbf{0.6}}$ & \textbf{97.8}$_{\pm \mathbf{0.5}}$ & \textbf{56.3}$_{\pm \mathbf{0.6}}$ \\
\bottomrule
\end{tabular}
} 
\vspace{-0.8em}
\end{table} 

\subsection{Preserving the Reverse-KL Objective Helps Mitigate Forgetting} 
\label{app:reverse_kl_forgetting} 

\begin{table}[ht]
    \centering
    \small
    \caption{Impact of preserving the reverse-KL objective on mitigating forgetting.}
    \label{tab:reverse_kl_forgetting}
    \resizebox{.5\linewidth}{!}{
    \begin{tabular}{@{}lcc@{}}
        \toprule
        \bf Method & \bf IFEval & \bf Arena-Hard \\
        \midrule
        \multicolumn{3}{c}{\cellcolor{gray!10}\textit{\textbf{Qwen3-0.6B}}} \\
        \midrule
        \textsc{Vanilla} & 60.1 & 8.6 \\
        \name $w/$ forward KL & 55.8\drop{4.3} & 7.2\drop{1.4} \\
        \name & \bf 64.3\up{4.2} & \bf 9.1\up{0.5} \\
        \midrule
        \multicolumn{3}{c}{\cellcolor{gray!10}\textit{\textbf{Qwen3-1.7B}}} \\
        \midrule
        \textsc{Vanilla} & 71.3 & 41.3 \\
        \name $w/$ forward KL & 65.7\drop{5.6} & 36.4\drop{4.9} \\
        \name & \bf 74.3\up{3.0} & \bf 43.6\up{2.3} \\
        \bottomrule
    \end{tabular}
    }
    \vspace{-1.0em}
\end{table}

Some distillation methods learn from teacher-generated tokens using a forward-KL objective, which may compromise previously acquired capabilities.
Beyond improving task performance~(\Cref{sec:ablation_study}), we investigate whether preserving the reverse-KL objective helps mitigate forgetting during distillation with teacher intervention.

We evaluate capability retention on IFEval~\citep{abs-2311-07911} and Arena-Hard~\citep{arenahard2024}, which are benchmarks designed to assess the subjective ability of LLMs.
We compare \name with a variant that replaces reverse KL with forward KL, using the original student checkpoint (\textsc{Vanilla}) as the reference for measuring changes in performance.

As shown in \Cref{tab:reverse_kl_forgetting}, the forward-KL variant underperforms the original checkpoint on both benchmarks at both student sizes, indicating degradation in previously acquired capabilities.
In contrast, \name improves over the original checkpoint across all four comparisons.
These results suggest that preserving the reverse-KL objective helps retain general capabilities while acquiring task-specific behaviors, providing an additional benefit beyond the agentic performance gains.

\subsection{Stochastic Intervention v.s. Deterministic Threshold} \label{app:stochastic_vs_deterministic} 

\paragraph{Setpup. }
To isolate the effect of stochastic intervention, we replace the proposed stochastic decision rule with deterministic thresholding while keeping all other training and evaluation settings unchanged.
Specifically, given the same turn-level discrepancy score $s_t$, the deterministic variant triggers teacher intervention whenever $s_t>\tau$, i.e.,
$z_t=\mathbf{1}[s_t>\tau]$.
We consider three thresholds, $\tau\in\{0.5,1.0,2.0\}$, covering increasingly conservative intervention policies.
All variants use the same student and teacher models, training data, optimization hyperparameters, rollout budget, and evaluation protocol as \name, so that the comparison isolates the effect of converting the discrepancy score into a stochastic intervention probability rather than a hard decision boundary.
We report results on the two long-horizon interactive benchmarks, ALFWorld and ScienceWorld, where compounding inter-turn errors make intervention decisions particularly consequential. 

\begin{table}[ht]
\centering 
\caption{Comparison of stochastic intervention and deterministic threshold on ALFWorld and ScienceWorld benchmarks.}
\label{tab:stochastic_vs_deterministic}
\resizebox{1.\linewidth}{!}{
\begin{tabular}{lccccccc}
\toprule
\multirow{2}{*}{\textbf{Method}} & \multicolumn{6}{c}{\cellcolor{cyan!20} \textbf{ALFWorld}} & \cellcolor{LimeGreen!20} \\
& \cellcolor{cyan!20} \textbf{Pick} 
& \cellcolor{cyan!20} \textbf{Pick Two} 
& \cellcolor{cyan!20} \textbf{Pick \& Clean} 
& \cellcolor{cyan!20} \textbf{Pick \& Heat} 
& \cellcolor{cyan!20} \textbf{Pick \& Cool} 
& \cellcolor{cyan!20} \textbf{Look} 
& \cellcolor{LimeGreen!20} \multirow{-2}{*}{\textbf{ScienceWorld}} \\
\midrule
Thre. $0.5$ & $88.5_{\pm 0.5}$ & $62.7_{\pm 0.4}$ & $36.5_{\pm 0.4}$ & $55.1_{\pm 0.5}$ & $49.7_{\pm 0.4}$ & $65.3_{\pm 0.5}$ & $20.8_{\pm 0.4}$ \\
Thre. $1.0$ & $89.8_{\pm 0.4}$ & $64.8_{\pm 0.4}$ & $38.0_{\pm 0.3}$ & $57.2_{\pm 0.5}$ & $51.8_{\pm 0.4}$ & $68.4_{\pm 0.4}$ & $23.2_{\pm 0.5}$ \\
Thre. $2.0$ & $89.0_{\pm 0.4}$ & $63.3_{\pm 0.5}$ & $35.8_{\pm 0.5}$ & $54.5_{\pm 0.6}$ & $49.2_{\pm 0.5}$ & $64.7_{\pm 0.6}$ & $20.3_{\pm 0.5}$ \\
\midrule
\name & \textbf{93.7}$_{\pm \mathbf{0.5}}$ & \textbf{69.1}$_{\pm \mathbf{0.5}}$ & \textbf{41.2}$_{\pm \mathbf{0.2}}$ & \textbf{62.0}$_{\pm \mathbf{0.6}}$ & \textbf{56.0}$_{\pm \mathbf{0.6}}$ & \textbf{73.7}$_{\pm \mathbf{0.5}}$ & \textbf{27.5}$_{\pm \mathbf{0.6}}$ \\
\bottomrule
\end{tabular}
} 
\end{table} 

\paragraph{Results and Analysis. }
\Cref{tab:stochastic_vs_deterministic} compares our stochastic intervention strategy with deterministic intervention using several fixed discrepancy thresholds.
Among the deterministic variants, a threshold of $1.0$ performs best across all tasks, while both smaller and larger thresholds lead to consistent degradation.
This non-monotonic behavior indicates that deterministic intervention is sensitive to the choice of decision boundary: a small threshold may trigger teacher intervention too frequently, whereas a large threshold may fail to intervene when corrective guidance is beneficial.
In contrast, \name consistently outperforms all deterministic variants across every ALFWorld category and ScienceWorld.
These results indicate that the performance of deterministic intervention depends noticeably on the selected threshold, whereas the stochastic strategy avoids introducing a manually tuned decision boundary and performs consistently better across the evaluated tasks.
By mapping discrepancy continuously to an intervention probability, \name allows teacher control to increase gradually with teacher--student disagreement while preserving opportunities for student execution.
This provides empirical support for using discrepancy-conditioned stochastic intervention rather than a fixed threshold.

\subsection{Teacher Computation Matched Comparison} \label{app:teacher_computation_matched} 

\paragraph{Setup.}
To examine whether the gains of \name arise simply from increased teacher usage, we compare all intervention-based methods under a matched teacher-intervention budget.
Specifically, we impose the same maximum number of teacher interventions for each method and terminate training once this budget is exhausted.
For methods with configurable intervention schedules or frequencies, we adjust these controls so that all methods consume the same total number of teacher interventions over training.
All other training and evaluation settings, including the student and teacher models, training data, optimization hyperparameters, rollout configuration, and evaluation protocol, are kept identical to the corresponding main experiments.
This comparison therefore controls for the amount of teacher intervention and focuses on how effectively each method allocates and learns from a fixed teacher budget.

\begin{table}[ht]
\centering 
\caption{Results of \name and baselines under the same teacher computation budget.}
\label{tab:teacher_computation_matched}
\resizebox{1.\linewidth}{!}{
\begin{tabular}{lccccccc}
\toprule
\multirow{2}{*}{\textbf{Method}} & \multicolumn{6}{c}{\cellcolor{cyan!20} \textbf{ALFWorld}} & \cellcolor{LimeGreen!20} \\
& \cellcolor{cyan!20} \textbf{Pick} 
& \cellcolor{cyan!20} \textbf{Pick Two} 
& \cellcolor{cyan!20} \textbf{Pick \& Clean} 
& \cellcolor{cyan!20} \textbf{Pick \& Heat} 
& \cellcolor{cyan!20} \textbf{Pick \& Cool} 
& \cellcolor{cyan!20} \textbf{Look} 
& \cellcolor{LimeGreen!20} \multirow{-2}{*}{\textbf{ScienceWorld}} \\
\midrule
\textsc{Relay-OPD} & $84.9_{\pm 0.5}$ & $44.8_{\pm 0.5}$ & $33.5_{\pm 0.4}$ & $32.4_{\pm 0.4}$ & $47.3_{\pm 0.5}$ & $35.1_{\pm 0.5}$ & $14.6_{\pm 0.4}$ \\
\textsc{Guided-OPD} & $86.2_{\pm 0.4}$ & $50.3_{\pm 0.5}$ & $34.8_{\pm 0.4}$ & $39.5_{\pm 0.4}$ & $48.2_{\pm 0.4}$ & $45.4_{\pm 0.5}$ & $13.1_{\pm 0.5}$ \\
\textsc{FutureBridge-OPD} & \underline{$88.3_{\pm 0.4}$} & \underline{$61.0_{\pm 0.5}$} & \underline{$36.7_{\pm 0.3}$} & \underline{$53.8_{\pm 0.4}$} & $49.7_{\pm 0.4}$ & \underline{$62.9_{\pm 0.5}$} & \underline{$18.7_{\pm 0.4}$} \\
\midrule
\name & \textbf{90.8}$_{\pm \mathbf{0.4}}$ & \textbf{65.8}$_{\pm \mathbf{0.4}}$ & \textbf{38.1}$_{\pm \mathbf{0.2}}$ & \textbf{58.6}$_{\pm \mathbf{0.5}}$ & \textbf{52.8}$_{\pm \mathbf{0.4}}$ & \textbf{69.9}$_{\pm \mathbf{0.4}}$ & \textbf{24.2}$_{\pm \mathbf{0.4}}$ \\
\bottomrule
\end{tabular}
}
\end{table} 

\paragraph{Results and Analysis.}
As shown in \Cref{tab:teacher_computation_matched}, \name consistently outperforms all competing intervention-based OPD methods under the same teacher-intervention budget.
Compared with the strongest baseline, \textsc{FutureBridge-OPD}, \name improves performance by $1.4$--$7.0$ points across all seven reported tasks, with an average gain of approximately $4.2$ points.
The largest improvements are observed on \textsc{Look} ($+7.0$), ScienceWorld ($+5.5$), and \textsc{Pick Two}/\textsc{Pick \& Heat} ($+4.8$ each), where successful completion requires sustained decision making over multiple interaction steps.
These results indicate that the advantage of \name cannot be explained solely by using more teacher interventions.
Instead, under a fixed teacher budget, its discrepancy-conditioned stochastic intervention allocates teacher guidance more effectively across interaction states, while the importance-weighted objective enables the student to learn from the resulting teacher-generated tokens without changing the conditional reverse-KL target.
Overall, the comparison suggests that the gains of \name arise from more effective use of teacher supervision rather than simply greater teacher involvement.

\subsection{Case Studies} 
\label{app:case_studies}

We pair early- and late-training trajectories from the same ALFWorld dataset instances. 
The controller column identifies the policy that generated the executed action, and $s_t$ is the turn-level discrepancy score of \Cref{eq:normalized_turn_score}. 
Scores on teacher-controlled turns are computed from the discarded student proposals. 
We abbreviate repeated search steps while preserving every action needed for the conclusions below.

\begin{case}{Local Transfer --- The Microwave Subtask Becomes Student-Controlled}{case_local_skill_transfer}
\begin{agetaskbox}
\textbf{Task:} Cool an apple and place it in a microwave.
\hfill \textbf{Dataset instance:} 15\\
\textbf{Early trajectory:} batch 1; 5 consecutive ineffective actions in the
shown segment.
\hfill \textbf{Later trajectory:} batch 112; 0 ineffective actions.\\
\textbf{Outcome:} Both trajectories succeed, but control of the previously problematic microwave subtask transfers from teacher to student.
\end{agetaskbox}

\vspace{0.4em}
\scriptsize
\setlength{\tabcolsep}{4pt}
\renewcommand{\arraystretch}{1.18}
\begin{tabularx}{\linewidth}{@{}l>{\raggedright\arraybackslash}Xcc@{}}
\toprule
\textbf{Turn} & \textbf{Executed action and environment response (abbreviated)}
& \textbf{Controller} & \textbf{$s_t$} \\
\midrule
\multicolumn{4}{@{}l}{\textit{Early training --- the student reaches an invalid local state and requires correction}} \\
15 & \texttt{cool apple 1 with fridge 1} $\rightarrow$ apple cooled
& Student & 0.993 \\
16--20 & Attempts to move/open the microwave $\rightarrow$
\texttt{Nothing happens.} five times
& Student & 0.478--0.835 \\
21 & \texttt{go to microwave 1} $\rightarrow$ arrives at closed microwave
& \cellcolor{ageTeacherBg}\textbf{Teacher} & 0.845 \\
22 & \texttt{open microwave 1} $\rightarrow$ microwave opened
& \cellcolor{ageTeacherBg}\textbf{Teacher} & 0.846 \\
23 & \texttt{move apple 1 to microwave 1} $\rightarrow$ reward 1
& Student & 0.506 \\
\midrule
\multicolumn{4}{@{}l}{\textit{Later training --- the same local subtask is completed without intervention}} \\
23 & \texttt{cool apple 1 with fridge 1} $\rightarrow$ apple cooled
& \cellcolor{ageStudentBg}\textbf{Student} & 0.286 \\
24 & \texttt{go to microwave 1} $\rightarrow$ arrives at closed microwave
& \cellcolor{ageStudentBg}\textbf{Student} & 0.231 \\
25 & \texttt{open microwave 1} $\rightarrow$ microwave opened
& \cellcolor{ageStudentBg}\textbf{Student} & 0.287 \\
26 & \texttt{move apple 1 to microwave 1} $\rightarrow$ reward 1
& \cellcolor{ageStudentBg}\textbf{Student} & 0.462 \\
\bottomrule
\end{tabularx}

\vspace{0.6em}
\begin{ageanalysisbox}
\footnotesize
Early in training, the student cools the apple but attempts to manipulate the microwave before navigating to it. 
The teacher restores a valid state by navigating to and opening the microwave. 
When the same instance reappears, the student independently executes this corrected sequence, and the navigation and opening scores decrease from $0.845/0.846$ to $0.231/0.287$. 
The case therefore provides evidence of \emph{local} skill acquisition: intervention is converted into student-controlled behavior on the corrected subtask. 
Because two teacher actions still occur earlier in the later trajectory, we do not interpret this example as end-to-end teacher independence.
\end{ageanalysisbox}
\end{case}

\begin{case}{End-to-End Transfer --- Search and Placement without Intervention}{case_search_place_transfer}
\begin{agetaskbox}
\textbf{Task:} Find two pens and place both in the desk.
\hfill \textbf{Dataset instance:} 713\\
\textbf{Early trajectory:} batch 23; 24 turns, 9 teacher interventions, and 4
ineffective actions.
\hfill \textbf{Later trajectory:} batch 134; 21 turns, 0 teacher interventions,
and 0 ineffective actions.\\
\textbf{Outcome:} Both trajectories receive reward 1 but the later trajectory is generated entirely by the student.
\end{agetaskbox}

\vspace{0.4em}
\scriptsize
\setlength{\tabcolsep}{4pt}
\renewcommand{\arraystretch}{1.18}
\begin{tabularx}{\linewidth}{@{}l>{\raggedright\arraybackslash}Xcc@{}}
\toprule
\textbf{Phase} & \textbf{Executed action sequence (abbreviated)}
& \textbf{Controller} & \textbf{$s_t$} \\
\midrule
\multicolumn{4}{@{}l}{\textit{Early training --- intervention spans exploration, object acquisition, and return navigation}} \\
Initialize search & \texttt{go drawer 1; open drawer 1; go drawer 2}
& \cellcolor{ageTeacherBg}\textbf{Teacher} & 0.292, 0.414, 0.698 \\
Locate first pen & Search shelves; \texttt{go shelf 5};
\texttt{take pen 1 from shelf 5}
& Mixed & 0.430 at intervention \\
Locate second pen & \texttt{go shelf 6; take pen 2 from shelf 6}
& \cellcolor{ageTeacherBg}\textbf{Teacher} & 0.454, 0.614 \\
Return and place & Teacher navigates to the desk; student eventually places
both pens $\rightarrow$ reward 1
& Mixed & 0.554, 0.288 at interventions \\
\midrule
\multicolumn{4}{@{}l}{\textit{Later training --- all search and placement decisions are student-generated}} \\
Initialize search & \texttt{go drawer 1; open drawer 1; go drawer 2; open drawer 2}
& \cellcolor{ageStudentBg}\textbf{Student} & 0.322--0.516 \\
Locate first pen & Search shelves; \texttt{go shelf 6; take pen 2 from shelf 6}
& \cellcolor{ageStudentBg}\textbf{Student} & 0.406--0.472 \\
Place first pen & \texttt{go desk 1; move pen 2 to desk 1}
& \cellcolor{ageStudentBg}\textbf{Student} & 0.161--0.357 \\
Locate and place second pen & \texttt{go shelf 5; take pen 1; go desk 1;
move pen 1 to desk 1} $\rightarrow$ reward 1
& \cellcolor{ageStudentBg}\textbf{Student} & 0.185--0.413 \\
\bottomrule
\end{tabularx}

\vspace{0.6em}
\begin{ageanalysisbox}
\footnotesize
This instance requires a longer dependency chain than Case~A: the agent must search multiple receptacles, remember two object locations, transport each object, and revisit the target desk. 
The early trajectory invokes the teacher at nine decisions distributed across these phases. 
In the later trajectory, the student reproduces the useful search and transport actions---including navigating to shelves 5 and 6, collecting both pens, and returning to the desk---without any teacher-generated turn. 
The simultaneous reductions in teacher interventions ($9\rightarrow0$), ineffective actions ($4\rightarrow0$), and trajectory length ($24\rightarrow21$) indicate that the improvement is not explained by longer exploration. 
Rather, the student executes the same multi-stage task autonomously and more efficiently.
\end{ageanalysisbox}
\end{case}

\begin{case}{Compositional Transfer --- The Full Skill Chain Becomes Student-Controlled}{case_compositional_transfer}
\begin{agetaskbox}
\textbf{Task:} Place a cooled bowl on the dining table.
\hfill \textbf{Dataset instance:} 176\\
\textbf{Early trajectory:} batch 6; 19 turns, 7 teacher interventions, and 2
ineffective actions.
\hfill \textbf{Later trajectory:} batch 117; 8 turns, 0 teacher interventions,
and 0 ineffective actions.\\
\textbf{Outcome:} Both trajectories receive reward 1 but the later student
executes the complete manipulation chain directly.
\end{agetaskbox}

\vspace{0.4em}
\scriptsize
\setlength{\tabcolsep}{4pt}
\renewcommand{\arraystretch}{1.18}
\begin{tabularx}{\linewidth}{@{}l>{\raggedright\arraybackslash}Xcc@{}}
\toprule
\textbf{Phase} & \textbf{Executed action sequence (abbreviated)}
& \textbf{Controller} & \textbf{$s_t$} \\
\midrule
\multicolumn{4}{@{}l}{\textit{Early training --- the student detours and the teacher supplies the critical manipulation chain}} \\
Acquire object & Student opens cabinet; teacher executes
\texttt{take bowl 1} and \texttt{go fridge 1}
& Mixed & 0.834, 0.754 at interventions \\
Detour & Student returns to the table, moves the bowl back, and repeatedly
examines the scene
& Student & 0.427--1.076 \\
Recover object & \texttt{take bowl 1 from diningtable 1}
& \cellcolor{ageTeacherBg}\textbf{Teacher} & 0.748 \\
Cool object & \texttt{open fridge 1; cool bowl 1 with fridge 1}
& \cellcolor{ageTeacherBg}\textbf{Teacher} & 0.417, 0.490 \\
Complete task & Teacher navigates to the table and finally executes
\texttt{move bowl 1 to diningtable 1} $\rightarrow$ reward 1
& Mixed & 0.601, 0.603 at interventions \\
\midrule
\multicolumn{4}{@{}l}{\textit{Later training --- the student executes a direct valid chain without intervention}} \\
Acquire object & \texttt{go cabinet 1; open cabinet 1; take bowl 1}
& \cellcolor{ageStudentBg}\textbf{Student} & 0.290--0.323 \\
Cool object & \texttt{go fridge 1; open fridge 1; cool bowl 1 with fridge 1}
& \cellcolor{ageStudentBg}\textbf{Student} & 0.308--0.491 \\
Complete task & \texttt{go diningtable 1; move bowl 1 to diningtable 1}
$\rightarrow$ reward 1
& \cellcolor{ageStudentBg}\textbf{Student} & 0.342--0.533 \\
\bottomrule
\end{tabularx}

\vspace{2.0em}
\begin{ageanalysisbox}
\footnotesize
The early trajectory exposes a compositional failure rather than a single
invalid action: after acquiring the bowl, the student returns it to the dining table before cooling it and then stalls on repeated examination. 
The teacher must re-acquire the object and supply the state-changing actions that open the fridge, cool the bowl, and complete the placement. 
On the later occurrence, the student executes precisely the required dependency order---acquire, navigate, open, cool, return, and place---in eight valid actions. 
Teacher use falls from seven turns to zero while trajectory length falls by more than half. 
Together with Case~\ref{cs:case_search_place_transfer}, this demonstrates full-trajectory student control in two distinct task families, complementing the local transfer illustrated by Case~\ref{cs:case_local_skill_transfer}.
\end{ageanalysisbox}
\end{case}

\section{Limitations and Future Work}
\label{app:limitations} 

While \name offers practical benefits, it also comes with a few limitations that point to interesting future work.
All experiments use the text-only LLMs, two student scales, two teacher scales, and two families of text-based interactive tasks. 
We do not test multimodal or embodied settings, or environments with irreversible external side effects. 
Whether discrepancy-calibrated intervention transfers to those settings is open.
The routing score in \Cref{eq:teacher_student_ratio} needs the teacher's per-token probabilities on student-proposed tokens, and the correction in \Cref{eq:importance_weighted_reverse_kl} needs them again on teacher-generated tokens. 
\name is therefore inapplicable to teachers exposed only through a sampling API, which excludes many of the strongest available models. 
When $z_t=1$ the teacher regenerates the entire turn. This preserves coherence between
the reasoning, the action, and the environment transition, but it spends teacher supervision on the whole response even when discrepancy is concentrated in a single decision, and it offers no mechanism for correcting part of a turn.

\end{document}